\documentclass[runningheads]{llncs}

\usepackage[T1]{fontenc}
\usepackage{graphicx}
\usepackage{hyperref}
\usepackage{color}

\usepackage{amsmath}
\usepackage{amssymb}
\usepackage{mathtools}
\usepackage[capitalize,noabbrev]{cleveref}
\usepackage{xcolor}
\usepackage{booktabs}
\usepackage{multirow}
\usepackage{makecell}
\usepackage{listings}
\usepackage{tikz}
\usepackage{pgfplots}
\pgfplotsset{compat=1.18}
\usepgfplotslibrary{colorbrewer,groupplots,fillbetween}
\usetikzlibrary{shapes, arrows.meta, positioning, calc, backgrounds, fit, shadows, decorations.markings, decorations.pathreplacing}
\usepackage[caption=false,font=footnotesize]{subfig} 

\definecolor{tabblue}{HTML}{1F77B4}
\definecolor{taborange}{HTML}{FF7F0E}
\definecolor{tabgreen}{HTML}{2CA02C}
\definecolor{tabred}{HTML}{D62728}
\definecolor{tabpurple}{HTML}{9467BD}
\definecolor{tabbrown}{HTML}{8C564B}
\definecolor{tabpink}{HTML}{E377C2}
\definecolor{tabgray}{HTML}{7F7F7F}
\definecolor{tabolive}{HTML}{BCBD22}
\definecolor{tabcyan}{HTML}{17BECF}

\definecolor{zoneBlue}{RGB}{235, 245, 255}
\definecolor{zoneGreen}{RGB}{235, 250, 235}
\definecolor{zonePurple}{RGB}{245, 235, 250}
\definecolor{zoneRed}{RGB}{255, 240, 240}
\definecolor{zoneOrange}{RGB}{255, 248, 235}
\definecolor{borderGray}{RGB}{180, 180, 180}
\definecolor{suggestionBlue}{RGB}{210, 235, 255}

\pgfplotscreateplotcyclelist{matplotlib}{
    {tabred},
    {tabgray},
    {taborange},
    {tabgreen},
    {tabblue},
    {tabpurple},
    {tabbrown},
    {tabpink},
    {tabolive},
    {tabcyan}
}

\definecolor{colorSupervised}{RGB}{0,0,0}        
\definecolor{colorSimCLR}{RGB}{230,159,0}        
\definecolor{colorBYOL}{RGB}{86,180,233}         
\definecolor{colorBarlow}{RGB}{102,67,57}        
\definecolor{colorVICReg}{RGB}{213,94,0}         
\definecolor{colorDINO}{RGB}{0,114,178}          
\definecolor{colorHypersolidNoL2}{RGB}{153,153,153}      
\definecolor{colorHypersolid}{RGB}{170,51,119}   
\definecolor{colorReSA}{RGB}{88,24,69}         
\definecolor{colorSwAV}{RGB}{128,200,128}        

\pgfplotscreateplotcyclelist{mycolorlist}{
    {colorSupervised, mark=*, mark size=3pt, thick},         
    {colorSimCLR, mark=square*, mark size=3pt},             
    {colorBYOL, mark=triangle*, mark size=3pt},             
    {colorSwAV, mark=triangle, mark size=3pt},
    {colorBarlow, mark=diamond*, mark size=3pt},            
    {colorVICReg, mark=o, mark size=3pt, thick},            
    {colorDINO, mark=square, mark size=3pt, thick},         
    {colorReSA, mark=star, mark size=3pt},
    {colorHypersolid, mark=pentagon*, mark size=3pt, thick}, 
    {colorHypersolidNoL2, mark=pentagon, mark size=3pt, thick}     
}

\definecolor{plasma0}{RGB}{13, 8, 135}
\definecolor{plasma1}{RGB}{34, 6, 144}
\definecolor{plasma2}{RGB}{49, 5, 151}
\definecolor{plasma3}{RGB}{63, 4, 156}
\definecolor{plasma4}{RGB}{78, 2, 162}
\definecolor{plasma5}{RGB}{91, 1, 165}
\definecolor{plasma6}{RGB}{103, 0, 168}
\definecolor{plasma7}{RGB}{116, 1, 168}
\definecolor{plasma8}{RGB}{129, 4, 167}
\definecolor{plasma9}{RGB}{141, 11, 165}
\definecolor{plasma10}{RGB}{152, 20, 160}
\definecolor{plasma11}{RGB}{162, 29, 154}
\definecolor{plasma12}{RGB}{173, 39, 147}
\definecolor{plasma13}{RGB}{182, 48, 139}
\definecolor{plasma14}{RGB}{191, 57, 132}
\definecolor{plasma15}{RGB}{199, 66, 124}
\definecolor{plasma16}{RGB}{207, 76, 116}
\definecolor{plasma17}{RGB}{214, 85, 109}
\definecolor{plasma18}{RGB}{221, 94, 102}
\definecolor{plasma19}{RGB}{227, 104, 95}
\definecolor{plasma20}{RGB}{233, 114, 87}
\definecolor{plasma21}{RGB}{239, 124, 81}
\definecolor{plasma22}{RGB}{243, 135, 74}
\definecolor{plasma23}{RGB}{247, 145, 67}
\definecolor{plasma24}{RGB}{250, 158, 59}
\definecolor{plasma25}{RGB}{252, 169, 52}
\definecolor{plasma26}{RGB}{253, 181, 46}
\definecolor{plasma27}{RGB}{253, 194, 41}
\definecolor{plasma28}{RGB}{252, 208, 37}
\definecolor{plasma29}{RGB}{249, 221, 37}
\definecolor{plasma30}{RGB}{245, 235, 39}
\definecolor{plasma31}{RGB}{240, 249, 33}

\pgfdeclareverticalshading{plasmabar}{100bp}{
  color(0bp)=(plasma0);
color(3bp)=(plasma1);
color(6bp)=(plasma2);
color(9bp)=(plasma3);
color(12bp)=(plasma4);
color(16bp)=(plasma5);
color(19bp)=(plasma6);
color(22bp)=(plasma7);
color(25bp)=(plasma8);
color(29bp)=(plasma9);
color(32bp)=(plasma10);
color(35bp)=(plasma11);
color(38bp)=(plasma12);
color(41bp)=(plasma13);
color(45bp)=(plasma14);
color(48bp)=(plasma15);
color(51bp)=(plasma16);
color(54bp)=(plasma17);
color(58bp)=(plasma18);
color(61bp)=(plasma19);
color(64bp)=(plasma20);
color(67bp)=(plasma21);
color(70bp)=(plasma22);
color(74bp)=(plasma23);
color(77bp)=(plasma24);
color(80bp)=(plasma25);
color(83bp)=(plasma26);
color(87bp)=(plasma27);
color(90bp)=(plasma28);
color(93bp)=(plasma29);
color(96bp)=(plasma30);
color(100bp)=(plasma31)
}

\newcommand{\plasmacolorbar}[2]{%
\begin{tikzpicture}[x=1pt,y=1pt,baseline=(current bounding box.center)]
  \shade[shading=plasmabar] (0,0) rectangle (#1,#2);
  \draw[line width=0.3pt] (0,0) rectangle (#1,#2);

  \draw[line width=0.3pt] (#1,0) -- (#1+3,0);
  \draw[line width=0.3pt] (#1,0.5*#2) -- (#1+3,0.5*#2);
  \draw[line width=0.3pt] (#1,#2) -- (#1+3,#2);

  \node[anchor=west,font=\footnotesize] at (#1+5,0) {0};
  \node[anchor=west,font=\footnotesize] at (#1+5,0.5*#2) {0.5};
  \node[anchor=west,font=\footnotesize] at (#1+5,#2) {1};
\end{tikzpicture}%
}

\definecolor{friendlykw}{HTML}{007020}
\definecolor{friendlycomment}{HTML}{60A0B0}
\definecolor{friendlystring}{HTML}{4070A0}
\definecolor{friendlynum}{HTML}{40A070}

\lstdefinestyle{mystyle}{
    commentstyle=\itshape\color{friendlycomment},
    keywordstyle=\bfseries\color{friendlykw},
    numberstyle=\tiny\color{gray},
    stringstyle=\color{friendlystring},
    basicstyle=\ttfamily\small,
    breakatwhitespace=false,
    breaklines=true,
    captionpos=b,
    keepspaces=true,
    numbers=left,
    numbersep=5pt,
    showspaces=false,
    showstringspaces=false,
    showtabs=false,
    tabsize=2,
    frame=none
}
\begin{document}

\title{Hypersolid: Emergent Vision Representations via Short-Range Repulsion}
\titlerunning{Hypersolid: Emergent Vision Representations via Short-Range Repulsion}

\author{Esteban Rodr\'iguez-Betancourt\inst{1}\orcidID{0009-0005-9002-6582} \and
Edgar Casasola-Murillo\inst{2}\orcidID{0000-0002-4863-1640}}

\authorrunning{E. Rodr\'iguez-Betancourt and E. Casasola-Murillo}

\institute{Posgrado en Computaci\'on e Inform\'atica,
Universidad de Costa Rica, San Jos\'e, Costa Rica\\
\email{esteban.rodriguezbetancourt@ucr.ac.cr} \and
Escuela de Ciencias de la Computaci\'on e Inform\'atica,
Universidad de Costa Rica, San Jos\'e, Costa Rica\\
\email{edgar.casasola@ucr.ac.cr}}

\maketitle

\begin{abstract}
A central problem in self-supervised learning is preventing representation collapse. Most methods avoid it through global mechanisms, such as contrastive expansion, variance constraints, decorrelating dimensions, or enforcing certain output distributions. In this work, we study a different design: short-range repulsion. We introduce Hypersolid, a self-supervised objective that combines view alignment with local collision avoidance.
Our method induces a latent geometry of compact, semantically aligned neighborhoods with low anisotropy. This geometry is especially effective for unsupervised clustering and fine-grained separation, although it comes at the cost of weaker transferability.

\keywords{Self-supervised learning \and representation learning}
\end{abstract}

\section{Introduction}

Most self-supervised learning methods optimize two complementary components: an \emph{alignment} objective that enforces consistency across views, and a \emph{separation} mechanism that prevents collapse. While there is broad consensus on alignment, separation strategies differ substantially, relying on approaches such as global expansion, redundancy reduction, or enforcing specific output distributions.

\begin{figure}[!htbp]
\begin{center}
\begin{tikzpicture}[
    font=\sffamily\footnotesize,
    >={Latex[width=1.8mm,length=1.8mm]},
    block/.style={
        rectangle, draw=black!70, fill=white, rounded corners=2pt,
        minimum height=0.6cm, align=center, drop shadow={opacity=0.15}, inner sep=2pt
    },
    vector/.style={
        rectangle, draw=black!80, fill=white, minimum width=0.5cm, minimum height=0.5cm,
        font=\bfseries\footnotesize, align=center
    },
    imgnode/.style={inner sep=0pt, drop shadow={opacity=0.2}, draw=white, line width=1pt},
    sum/.style={circle, draw=black, fill=white, inner sep=1.5pt},
    arrow/.style={->, thick, draw=black!70},
    dashed_arrow/.style={->, dashed, thick, draw=black!60},
    bus/.style={thick, draw=black!60, rounded corners=3pt},
]

    \node[imgnode, label={[font=\sffamily\footnotesize\bfseries]below:Input $X$}] (input) at (0,0) {
        \includegraphics[width=1.0cm]{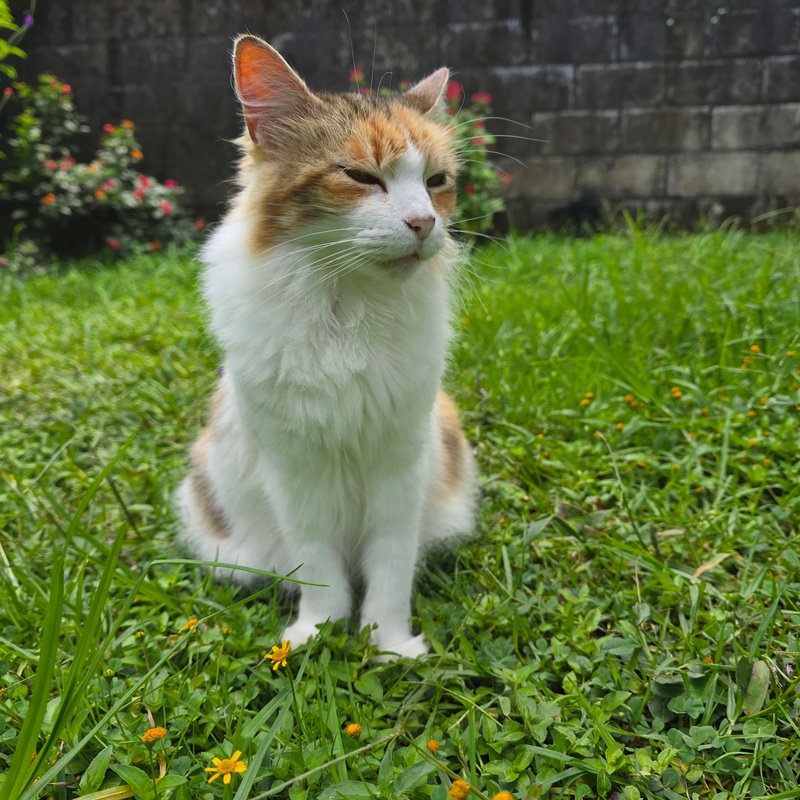}
    };

    \node[imgnode] (aug2) at (1.35, 0) {
        \includegraphics[width=0.6cm, trim={100 100 400 400}, clip]{images_demo/cat1.jpg}
    };
    \node[imgnode] (aug1) at (1.35, 0.9) {
        \includegraphics[width=0.6cm, trim={150 200 100 100}, clip]{images_demo/cat1.jpg}
    };
    \node[imgnode] (aug3) at (1.35, -1.2) {
        \includegraphics[width=0.6cm, trim={200 400 200 0}, clip]{images_demo/cat1.jpg}
    };
    \coordinate (augtop) at (1.35, 1.5);

    \draw[arrow] (input.east) -- (aug1.west);
    \draw[arrow] (input.east) -- (aug2.west);
    \draw[arrow] (input.east) -- (aug3.west);

    \node[block, fill=green!10, text width=1.0cm, minimum height=4.0cm] (encoder) at (2.85, 0) {$f_\theta$};

    \draw[arrow] (aug1.east) -- (encoder.west |- aug1.east);
    \draw[arrow] (aug2.east) -- (encoder.west |- aug2.east);
    \draw[arrow] (aug3.east) -- (encoder.west |- aug3.east);

    \node[vector, anchor=west] (z1) at ([xshift=0.1cm, yshift=1.2cm]encoder.east) {$z_1$};
    \node[vector, anchor=west] (z2) at ([xshift=0.1cm]encoder.east) {$z_2$};
    \node[vector, anchor=west] (z3) at ([xshift=0.1cm, yshift=-1.2cm]encoder.east) {$z_3$};

    \draw[thick, black!70] (encoder.east |- aug1.east) -- (z1.west);
    \draw[thick, black!70] (encoder.east |- aug2.east) -- (z2.west);
    \draw[thick, black!70] (encoder.east |- aug3.east) -- (z3.west);

    \coordinate (LossLeft) at (4.5, 0);
    \coordinate (LossRight) at (10.2, 0);

    \coordinate (AlignY) at (0, 1.5);
    \node[block, fill=violet!5] (maxpool) at ([xshift=1.5cm]LossLeft |- AlignY) {Max Pool};
    \node[block, fill=violet!5, right=0.7cm of maxpool] (cosloss) {Cosine\\Distance};

    \draw[arrow] (maxpool) -- node[midway, inner sep=1pt, font=\bfseries\footnotesize] {\textbackslash{}\kern-0.15em\textbackslash{}} (cosloss);

    \coordinate (RepY) at (0, 0);
    \node[block, fill=red!5, text width=2.4cm] (pairwise) at ([xshift=1.5cm]LossLeft |- RepY) {Pairwise Cosine\\Similarity};
    \node[block, fill=red!5, right=0.5cm of pairwise] (calc) {$\frac{\text{ReLU}(sim - \alpha)}{1-\alpha}$};

    \draw[arrow] (pairwise) -- (calc);

    \coordinate (NormY) at (0, -1.7);
    \node[block, fill=orange!5, text width=2.0cm] (norm) at ([xshift=1.5cm]LossLeft |- NormY) {$L_2$ Penalty};

    \coordinate (BusMerge) at (4.25, 0);

    \draw[bus] (z1.east) -- (BusMerge |- z1.east) -- (BusMerge |- z3.east) -- (z3.east);
    \draw[bus] (z2.east) -- (BusMerge |- z2.east);

    \draw[arrow] (BusMerge |- z1.east) -- (maxpool.west);

    \draw[bus, ->] (BusMerge |- z1.east) to[out=0, in=-160] (cosloss.south);

    \draw[arrow] (BusMerge |- z2.east) -- (pairwise.west);

    \coordinate (BranchPoint) at (4.5, -0.5);
    \draw[dashed_arrow] (BusMerge |- z2.east) -- (BranchPoint) -- node[below, font=\scriptsize, align=center]{All other images} (pairwise.south west);

    \draw[arrow] (BusMerge |- z3.east) -- (norm.west);

    \node[sum] (sigma) at (10.5, 0) {$\sum$};

    \draw[arrow] (cosloss.east) -| (sigma.north);
    \draw[arrow] (calc.east) -- (sigma.west);
    \draw[arrow] (norm.east) -| (sigma.south);

    \begin{pgfonlayer}{background}
        \node[rounded corners, draw=borderGray, fill=zoneBlue, fit=(input)(aug1)(aug3)(augtop), inner sep=8pt,
              label={[anchor=north west, inner sep=2pt, font=\sffamily\footnotesize\bfseries]north west:Augmentation}] {};

        \path (encoder.north) ++(0, 8pt) coordinate (EncTop);
        \node[rounded corners, draw=borderGray, fill=zoneGreen, inner sep=5pt, fit=(encoder)(z1)(z3)(EncTop),
              label={[anchor=north, inner sep=2pt, font=\sffamily\footnotesize\bfseries]north:Encoder}] {};

        \draw[rounded corners, draw=borderGray, fill=zonePurple]
             (LossLeft |- 2.5, 1.0) rectangle (LossRight |- 2.5, 2.5);
        \node[font=\sffamily\footnotesize\bfseries, anchor=north west] at (LossLeft |- 2.5, 2.5) {Alignment};

        \draw[rounded corners, draw=borderGray, fill=zoneRed]
             (LossLeft |- -0.7, -0.7) rectangle (LossRight |- 0.85, 0.85);
        \node[font=\sffamily\footnotesize\bfseries, anchor=north west] at (LossLeft |- -0.7, 0.85) {Hard-Ball Repulsion};

        \draw[rounded corners, draw=borderGray, fill=zoneOrange]
             (LossLeft |- -2.0, -0.9) rectangle (LossRight |- -2.0, -2.25);
        \node[font=\sffamily\footnotesize\bfseries, anchor=north west] at (LossLeft |- -2.0, -0.9) {Normalization};

    \end{pgfonlayer}

\end{tikzpicture}
\caption{\textbf{Hypersolid Training Workflow.} An input image is augmented into global and local views and encoded. \textbf{Top (Purple):} Views are aligned to a ``Feature Union'' target created by max-pooling embeddings (with stop-gradient). \textbf{Middle (Red):} Short-range repulsion penalizes any pair (positive or negative) exceeding similarity $\alpha$. \textbf{Bottom (Yellow):} A weak $L_2$ penalty regularizes feature magnitude.}
\label{fig:method_workflow}
\end{center}
\end{figure}

Instead of using a mechanism that relies on global properties of the learned latent space, \emph{Hypersolid} leverages short-range repulsion constraints to avoid collapse. Our proposed method, summarized in \cref{fig:method_workflow}, repels all embedding pairs within a short-range exclusion zone (including positive pairs), attracts each augmentation toward the max-pooling of its embeddings, and adds a weak L2 regularization. We found the resulting latent space to be globally isotropic and sparse, with tighter intra-class neighborhoods and larger inter-cluster margins. This geometry is particularly effective for automatic clustering and fine-grained separation. However, compared with general-purpose SSL methods, the features learned by Hypersolid are less transferable across broad downstream tasks.

\section{Related Work}\label{sec:relatedwork}

\subsubsection{Contrastive Learning and Global Expansion.}
Methods like SimCLR~\cite{chen2020simclr} and MoCo~\cite{he2020moco} enforce global repulsion, treating every image as a distinct class and pushing all negative pairs apart, thereby aiming to expand the representation distribution to fill the latent space. In contrast, Hypersolid enforces a universal exclusion zone for all pairs. While positive pairs are aligned, they are  repelled if they get into this short-range radius, enforcing a minimum separation distance even between views from the same image.

\subsubsection{Non-Contrastive Regularization.}
Barlow Twins~\cite{Zbontar2021} and VICReg~\cite{bardes2022vicreg} minimize feature redundancy or enforce variance constraints, maximizing a proxy for differential entropy. Hypersolid does not directly optimize toward maximizing differential entropy; instead, it simply uses short-range repulsion to avoid collapse locally, assigning zero gradient force toward separating elements outside this region. Surprisingly, the resulting geometry self-organized into a latent space that is globally isotropic.
Other non-contrastive Siamese methods use architectural asymmetry or stop-gradient to avoid collapse. BYOL~\cite{grill2020bootstrap} uses an online predictor and a momentum target encoder, whereas SimSiam~\cite{chen2021simsiam} shows that a stop-gradient operation and a predictor can suffice without negative pairs, a momentum encoder, or large batches. These methods primarily enforce invariance between views; Hypersolid additionally preserves local distinctions through its repulsion term.

\subsubsection{Clustering and Discretization.} Clustering approaches such as SwAV explicitly discretize the space via prototypes. Hypersolid shares a similar intuition regarding discretization but, rather than collapsing samples into shared prototypes, it attempts to distinguish every single augmentation. In this regard, the tight clusters observed in Hypersolid are an emergent result of the training dynamics, not something explicitly enforced in the model.
ReSA~\cite{weng2025resa} is a more recent clustering-oriented SSL method: it uses the clustering structure already present in the encoder representations as a positive feedback signal during training. Unlike ReSA, Hypersolid does not assign samples to prototypes or use cluster assignments in its training loss; its clusters emerge from the interaction between alignment and short-range exclusion.

\subsubsection{Short Range Repulsion and Packing.}
Both Hypersolid and CLAMP \cite{zhang2025contrastiveselfsupervisedlearningneural} use short-range repulsion as a core mechanism. However, both methods differ in how this repulsion is applied. CLAMP explicitly groups augmentations of each image into a sub-manifold. It performs packing at the sub-manifold level, as it penalizes overlaps between different sub-manifolds of different images, while it simultaneously attempts to contract the sub-manifolds of the same image. In contrast, Hypersolid performs the packing at the individual embedding level, as it applies a short-range repulsion between any pair of embeddings that get closer than a fixed threshold, even if both augmentations belong to the same image. Another difference is that Hypersolid does not attempt to contract the sub-manifolds of the same image; instead, it builds a target embedding using a max-pool operation, and aligns all augmentations of the same image towards this new target.

\section{Method Description}\label{sec:Description}

Our method follows a structure similar to other self-supervised learning methods. We use a neural network to produce embeddings for different views of an image, then use those embeddings to build a target, and finally train the network to follow those targets according to our loss function.

For producing the views, we opted to use the same recipe as DINO~\cite{caron2021dino}: at least two global views and $N$ local views. Other than preserving a different size of the original image, global and local views are treated equally. The full list of augmentations is described in \cref{sec:APTrainingAugmentations}.

Our loss function can be expressed as
\[
    \mathcal{L}_\text{Hypersolid} = \mathcal{L}_\text{alignment} + \mathcal{L}_\text{repulsion} + \mathcal{L}_\text{normalization}
\]
where $\mathcal{L}_\text{alignment}$ encourages views to have similar representations, $\mathcal{L}_\text{repulsion}$ strongly penalizes representations that get too close, and finally $\mathcal{L}_\text{normalization}$ applies a weak $L_2$ penalty, which makes the representations sparser.

\subsection{Repulsion: Enforce Short-Range Separation}
We enforce a maximum cosine similarity $\alpha$ between all view embeddings (positive and negative). Crucially, gradients are zero for pairs with similarity below $\alpha$. Formally, let $\mathcal{Z}=\{z_1, z_2, \ldots, z_M\}$ be the set of all embeddings in the batch, where $M = B \times V$ ($B$ images, each with $V$ views). Then, the repulsion loss is expressed as:

\[
\mathcal{L}_\text{repulsion}
= \mathbb{E}_{z_i, z_j \in \mathcal{Z}} \left[ \frac{\mathrm{ReLU}(\cos(z_i,z_j) - \alpha)}{1 - \alpha} \right]
\]

where the expectation is taken over all pairs in the batch.

This formulation represents a deliberate deviation from the standard objective of learning invariance under augmentation. By applying repulsion even to positive pairs (views of the same image), we enforce a minimum degree of separation between augmentations, effectively encouraging the model to preserve local variation among augmentations while still aligning them through the target.

The ReLU function creates a sparse gradient landscape, preventing the optimizer from wasting updates on separating representations that already satisfy the exclusion threshold. Repulsion dominates only within the exclusion zone ($\text{similarity} > \alpha$), ensuring geometric constraints take precedence over alignment only when necessary.

\subsection{Alignment: Feature Union via Max-Pooling}
For the alignment term, we construct a target embedding for each source image by taking the dimension-wise maximum of the representations across its augmentations. Let $\mathcal{V} \subset \mathcal{Z}$ be the subset of embeddings corresponding to a single image. We define its target representation, $z_\text{target}$, as:
\[
    z^{(k)}_\text{target} = \max_{z \in \mathcal{V}}(z^{(k)})
\]

where $k$ indexes the feature dimension. This target acts as a bag of features, representing a union of all salient features visible across the augmentations. Empirically, we found that this max-pooling strategy prevented the early optimization stagnation observed when using a mean centroid.

Additionally, we apply a stop-gradient operation to the target. Without this, the optimization ends up reducing the magnitude of the view embeddings, rather than fully aligning them with the target, which causes early learning stagnation. The final alignment loss minimizes the cosine distance between each normalized view $z_i$ and the normalized target $z_\text{target}$.

\subsection{$L_2$ Embedding Normalization}

Finally, we apply a weak penalty to the $L_2$ norm of the pre-normalized embeddings. Formally:
\[
\mathcal{L}_\text{normalization} = \lambda \left(\mathbb{E}_z[\Vert z \Vert_2] - 1\right)^2
\]

where the expectation is taken over all the views in the batch. This term serves two functions. First, it caps the growth of feature magnitudes, ensuring that the max-pooling target selects features based on semantic activation rather than arbitrary scale. By preventing specific dimensions from dominating purely due to unbounded growth, this constraint actively mitigates representation anisotropy.

Second, the weight $\lambda$ controls the optimization trajectory. We found that a weak penalty (such as $\lambda=10^{-6}$) behaves better than higher values. We hypothesize that a weaker penalty accelerates convergence by guiding the shape of the latent space, without forcing the embeddings to move along a ``geodesic route.''

As will be analyzed in the following sections, the L2 regularization causes measurable effects in how tight clusters are, the distribution of pairwise similarities, the number of clusters produced, etc.

\subsection{Method implementation}\label{subsec:method_implementation}
The full code of our loss function is presented as follows:

\begin{lstlisting}[language=Python]
class HypersolidLoss(nn.Module):
  def __init__(self, alpha=0.9, norm_factor=1e-6):
    super().__init__()
    self.alpha = alpha
    self.norm_factor = norm_factor

  def forward(self, feats):
    B, V, D = feats.shape
    x = F.normalize(feats, dim=-1)   # [B, V, D]

    # Alignment term:
    dst, _ = feats.max(dim=1, keepdim=True) # [B, 1, D]
    c = F.normalize(dst, dim=-1).detach() # [B, 1, D]
    pos_sim = x @ c.transpose(1, 2)  # [B, V, 1]
    pos_sim = pos_sim.squeeze(2)     # [B, V]

    alignment = (1 - pos_sim).mean()

    # Pairwise Repulsion term:
    all_feats = x.reshape(B*V, D)    # [B*V, D]
    sim = all_feats @ all_feats.T    # [B*V, B*V]
    sim.fill_diagonal_(0.0) # remove self-similarity

    repulsion = F.relu(sim - self.alpha) / (1 - self.alpha)
    repulsion = repulsion.mean()

    # Normalization term
    l2norm = (feats.norm(p=2, dim=-1).mean() - 1)**2
    l2norm = l2norm * self.norm_factor

    return alignment + repulsion + l2norm
\end{lstlisting}

\section{Methodology}\label{sec:methodology}

We now describe our evaluation methodology. In the following subsections, we detail the datasets, backbones, training protocol, evaluation metrics, and data augmentation strategy used throughout this work.

\subsection{Datasets and Backbones}

We evaluate Hypersolid on a diverse set of image classification benchmarks: STL‑10~\cite{coates2011stl10}, CIFAR‑10 and CIFAR‑100~\cite{krizhevsky2009learning}, Food‑101~\cite{bossard14}, ImageNet‑100, and ImageNet‑1k~\cite{deng2009imagenet}. In the case of STL-10, CIFAR-10, CIFAR-100 and Food-101 we trained a ResNet-18~\cite{he2016resnet} backbone for 200 epochs. For ImageNet-100, we trained a ResNet-18~\cite{he2016resnet} for 200 epochs with Hypersolid (with L2 normalization), and compared against published SSL benchmarks. For ImageNet-1k, we used a ResNet-50~\cite{he2016resnet} trained for 100 epochs, following the same protocol as the baseline methods.

\subsection{Training Protocol}

All models trained by us, unless noted otherwise, were trained with the AdamW optimizer~\cite{loshchilov2019adamw} and a base learning rate of $10^{-3}$, using BF16 mixed precision on a single NVIDIA H100 GPU. For STL‑10, CIFAR‑10, CIFAR‑100, and Food‑101, we set the batch size to 128. For ImageNet‑1k, Hypersolid and the supervised baseline were trained with a batch size of 512; other SSL baselines used their respective optimal configurations: ReSA with batch size 256 (authors' checkpoint), DINO with batch size 128 (Lightly checkpoint), and SimCLR, BYOL, Barlow Twins, VICReg, and SwAV with batch size 256 (Lightly checkpoints). All ImageNet‑1k models were trained for 100 epochs.

The default Hypersolid configuration uses a repulsion threshold $\alpha = 0.9$, a projector dimension of 1024, 2 global views and 6 local views (augmentation details in Section~\ref{sec:APTrainingAugmentations}), and an $L_2$ normalization weight $\lambda = 10^{-6}$ (see Section~\ref{sec:Description}). To disable the $L_2$ normalization in Hypersolid, we set $\lambda = 0$.

\subsection{Evaluation Metrics}

We measure representation quality through linear evaluation and $k$-nearest neighbors (KNN) classification. For linear evaluation, we trained a linear probe classifier on top of the frozen backbone features using the official training set and report top‑1 accuracy on the validation/test set, following the same setup as \cite{kalapos2024whiteningconsistentlyimprovesselfsupervised}. For KNN, we compute the top‑1 accuracy using cosine distance between test and training embeddings, with $K$ chosen per dataset: $K=200$ for ImageNet‑1k and Food‑101, $K=20$ for ImageNet‑100, and $K=5$ for STL‑10, CIFAR‑10, and CIFAR‑100. Clustering quality is measured with normalized mutual information (NMI) and adjusted Rand index (ARI) against the ground‑truth labels, using K‑Means ($k=1000$) and density‑based algorithms (DBSCAN, HDBSCAN) as described in Section \ref{subsec:clustering_properties}.

\subsection{Training Augmentations}\label{sec:APTrainingAugmentations}

For our augmentation pipeline, we leveraged the Lightly Framework transforms for DINO. The parameter values are described in \cref{tab:augmentations}. For ImageNet-1k and Food-101, we used the default \texttt{DINOTransform} settings for both Hypersolid and DINO. The selected values were not fine-tuned; their choice was based on dataset image sizes and network constraints.

\begin{table}[!htbp]
\caption{Augmentations used in Hypersolid.}
\label{tab:augmentations}
\begin{center}
\small
\begin{tabular}{lll}
\toprule
Parameter & Library Default & Changes \\
\midrule
Global Crop Size & 224 & CIFAR: 32,\; STL-10: 96 \\
Global Crop Scale & (0.4, 1.0) & CIFAR: (0.8, 1.0) \\
Local Crop Size & 96 & CIFAR: 32,\; STL-10: 48,\; ViT: 224 \\
Local Crop Scale & (0.05, 0.4) & CIFAR: (0.08, 0.9) \\
Number of Local Views & 6 &  \\
Horizontal Flip Prob. & 0.5 &  \\
Vertical Flip Prob. & 0 &  \\
Random Rotation Prob. & 0 &  \\
Random Rotation Degrees & None &  \\
Color Jitter Prob. & 0.8 &  \\
Color Jitter Strength & 0.5 &  \\
Brightness Jitter & 0.8 &  \\
Contrast Jitter & 0.8 &  \\
Saturation Jitter & 0.4 &  \\
Hue Jitter & 0.2 &  \\
Random Grayscale Prob. & 0.2 &  \\
Gaussian Blur & (1.0, 0.1, 0.5) & CIFAR: Disabled \\
Gaussian Blur Sigmas & (0.1, 2) &  \\
Kernel Size & None &  \\
Kernel Scale & None &  \\
Solarization Prob. & 0.2 &  \\
Normalization & ImageNet mean/std &  \\
\bottomrule
\end{tabular}
\end{center}
\end{table}

\section{Results}\label{sec:MainResults}

In this section, we examine the performance and geometric properties of representations learned by Hypersolid. First, we report classification accuracy under linear probe and k-NN evaluation, then analyze the global geometry, pairwise similarities, and clustering quality of the latent space, followed by a qualitative visualization of the learned features.

\subsection{Classification Accuracy}
\Cref{tab:accuracy-table} reports Top-1 accuracy for Linear Probe and k-NN classifiers. For the k-NN classifier we used different values of $K$: $K=200$ for ImageNet-1k and Food-101, $K=20$ for ImageNet-100, and $K=5$ for STL-10, CIFAR-10, and CIFAR-100. Hypersolid achieved the best performance on CIFAR-10, CIFAR-100, ImageNet-100, and Food-101, while remaining competitive on STL-10. This suggests that Hypersolid is particularly well suited to low-resolution or fine-grained datasets; however, it starts lagging on datasets that require highly descriptive features, such as ImageNet-1k.

\begin{table}[!htbp]
\caption{Linear probe and KNN accuracy.}
\label{tab:accuracy-table}
\begin{center}
\small
\setlength{\tabcolsep}{2.5pt}
\begin{tabular}{@{}c@{\hspace{0.8em}}c@{}}

\begin{tabular}[t]{@{}lrr@{}}
\toprule
\bfseries Method & \bfseries Linear & \bfseries K-NN \\

\midrule
\multicolumn{3}{@{}l}{\bfseries CIFAR-10} \\
\textcolor{gray}{Supervised} & \textcolor{gray}{84.17} & \textcolor{gray}{84.16} \\
SimCLR & 72.58 & 69.16 \\
BYOL & 69.37 & 65.54 \\
Barlow Twins & 74.63 & 71.81 \\
VICReg & 74.72 & 72.67 \\
DINO & 72.46 & 70.07 \\
Hypersolid & \textbf{83.91} & \textbf{82.68} \\

\midrule

\multicolumn{3}{@{}l}{\bfseries STL-10} \\
\textcolor{gray}{Supervised} & \textcolor{gray}{71.86} & \textcolor{gray}{72.43} \\
SimCLR & \textbf{82.89} & \textbf{79.16} \\
BYOL & 81.28 & 77.14 \\
Barlow Twins & 82.19 & 78.20 \\
VICReg & 81.69 & 78.48 \\
DINO & 79.75 & 76.85 \\
Hypersolid & 82.11 & 77.60 \\

\midrule
\multicolumn{3}{@{}l}{\bfseries ImageNet-100$^*$} \\
Barlow Twins & - & 46.5 \\
BYOL         & - & 43.9 \\
DINO         & - & 51.8 \\
FastSiam     & - & 55.9 \\
MoCo         & - & 56.0 \\
NNCLR        & - & 45.3 \\
SimCLR       & - & 46.9 \\
SimSiam      & - & 53.4 \\
SwAV         & - & 67.8 \\
Hypersolid   & 66.76 & \textbf{68.4} \\

\bottomrule
\end{tabular}

&

\begin{tabular}[t]{@{}lrr@{}}
\toprule
\bfseries Method & \bfseries Linear & \bfseries K-NN \\

\midrule
\multicolumn{3}{@{}l}{\bfseries CIFAR-100} \\
\textcolor{gray}{Supervised} & \textcolor{gray}{53.05} & \textcolor{gray}{51.19} \\
SimCLR & 39.93 & 32.92 \\
BYOL & 38.68 & 33.91 \\
Barlow Twins & 41.52 & 36.46 \\
VICReg & 43.31 & 37.55 \\
DINO & 39.27 & 34.07 \\
Hypersolid & \textbf{55.06} & \textbf{51.92} \\

\midrule

\multicolumn{3}{@{}l}{\bfseries Food-101} \\
\textcolor{gray}{Supervised} & \textcolor{gray}{71.33} & \textcolor{gray}{70.01} \\
SimCLR & 61.15 & 49.99 \\
BYOL & 58.21 & 45.83 \\
Barlow Twins & 64.11 & 54.90 \\
VICReg & 65.69 & 56.61 \\
DINO & 64.48 & 55.43 \\
Hypersolid & \textbf{71.32} & \textbf{64.27} \\

\midrule

\multicolumn{3}{@{}l}{\bfseries ImageNet-1k$^\dagger$} \\
\textcolor{gray}{Supervised} & \textcolor{gray}{70.02} & \textcolor{gray}{61.72} \\
SimCLR & 62.03 & 45.60 \\
BYOL & 61.84 & 45.88 \\
SwAV & \textbf{66.15} & 50.19 \\
Barlow Twins & 59.69 & 46.24 \\
VICReg & 62.66 & 47.14 \\
DINO & 64.43 & 51.28 \\
ReSA & 60.85 & \textbf{60.79} \\
Hypersolid & 55.59 & 51.14 \\
--- w/o L2 & 60.38 & 47.03 \\

\bottomrule
\end{tabular}

\end{tabular}

{\footnotesize\raggedright
$^*$ KNN ($K=20$, 200 epochs) reported by Lightly Framework benchmarks.

$^\dagger$ Linear probe and KNN ($K=200$) evaluated by us using standard center-crop protocol.
\par}

\end{center}
\end{table}

\subsection{Geometric Properties}\label{sec:geometryProperties}

\begin{table}[tbp]
\caption{Geometric Properties of Learned Representations on ImageNet-1k (ResNet-50) and Food-101 (ResNet-18). Abbreviations: Aniso.\ (Anisotropy), Corr.\ (Feature Correlation), CVN (Center Vector Norm), MPA (Mean Pairwise Angle), and $d^\prime$ (Sensitivity Index).}
\label{tab:qualityMetrics}
\begin{center}
\begin{tabular}{clcccccccc}
\toprule
& Method & Aniso. & Corr. & CVN & \makecell{Centroid\\Rank} & \makecell{Embed.\\Rank} & Sparsity & $d^\prime$ & MPA \\
\midrule
\multirow{10}{*}{\rotatebox[origin=c]{90}{\bfseries ImageNet-1k}}
& \textcolor{gray}{Supervised} & \textcolor{gray}{0.29} & \textcolor{gray}{0.043} & \textcolor{gray}{0.53} & \textcolor{gray}{499} & \textcolor{gray}{1656} & \textcolor{gray}{21.91\%} & \textcolor{gray}{1.70} & \textcolor{gray}{$73.60^\circ$} \\
& SimCLR & 0.26 & 0.059 & 0.52 & 407 & 1372 & 32.94\% & 1.69 & $74.42^\circ$ \\
& BYOL & 0.35 & 0.078 & 0.69 & 317 & 1085 & 7.69\% & 1.67 & $61.02^\circ$ \\
& SwAV & 0.24 & 0.049 & 0.51 & 445 & 1396 & 26.87\% & 1.86 & $74.55^\circ$ \\
& Barlow Twins & 0.39 & 0.071 & 0.63 & 350 & 1292 & 19.85\% & 1.70 & $66.40^\circ$ \\
& DINO & 0.53 & 0.076 & 0.74 & 308 & 933 & 0.50\% & 2.04 & $56.18^\circ$ \\
& VICReg & 0.25 & 0.056 & 0.51 & 458 & 1387 & 41.69\% & 1.84 & $74.86^\circ$ \\
& ReSA & 0.21 & 0.041 & 0.46 & \textbf{532} & \textbf{1553} & 41.61\% & 2.00 & $77.61^\circ$ \\
& Hypersolid & \textbf{0.11} & \textbf{0.036} & \textbf{0.33} & 452 & 1013 & \textbf{71.79\%} & 2.16 & $\mathbf{83.52^\circ}$ \\
& --- w/o L2 & 0.32 & 0.094 & 0.55 & 329 & 800 & 28.18\% & \textbf{2.32} & $71.97^\circ$ \\
\midrule
\multirow{7}{*}{\rotatebox[origin=c]{90}{\bfseries Food-101}}
& \textcolor{gray}{Supervised} & \textcolor{gray}{0.49} & \textcolor{gray}{0.075} & \textcolor{gray}{0.69} & \textcolor{gray}{52} & \textcolor{gray}{372} & \textcolor{gray}{5.41\%} & \textcolor{gray}{1.42} & \textcolor{gray}{$61.20^\circ$} \\
& SimCLR & 0.41 & 0.087 & 0.65 & 45 & 330 & 21.18\% & 1.05 & $64.54^\circ$ \\
& BYOL & 0.71 & 0.109 & 0.84 & 25 & 251 & 1.85\% & 0.76 & $44.47^\circ$ \\
& Barlow Twins & 0.21 & 0.077 & \textbf{0.48} & 63 & 359 & \textbf{41.33\%} & 1.28 & $76.40^\circ$ \\
& DINO & 0.28 & \textbf{0.064} & 0.54 & 56 & \textbf{392} & 25.54\% & 1.21 & $72.71^\circ$ \\
& VICReg & 0.28 & 0.079 & 0.55 & 61 & 358 & 32.65\% & 1.27 & $72.48^\circ$ \\
& Hypersolid & \textbf{0.19} & 0.074 & \textbf{0.48} & \textbf{68} & 290 & 38.16\% & \textbf{1.90} & $\mathbf{76.52^\circ}$ \\
\bottomrule
\end{tabular}
\end{center}
\end{table}

To better understand how the short‑range repulsion shapes the latent space, we examine several quantitative properties of the learned embeddings: anisotropy, feature correlation, the effective rank of centroids and embeddings, sparsity, sensitivity index (\(d^\prime\)), and mean pairwise angle.

Let $X\in\mathbb{R}^{N\times D}$ denote the matrix of evaluated embeddings, with rows $x_i$, where $N$ is the number of samples, $D$ is the embedding dimension, and $X_{ik}$ denotes the $k$-th coordinate of $x_i$.

\paragraph{Anisotropy.}
We define anisotropy as the squared spectral concentration of $X$. If $\sigma_k$ are the singular values of $X$, then
\[
    \operatorname{Aniso}(X)=\frac{\sigma_1^2}{\sum_k\sigma_k^2}.
\]

\paragraph{Center-vector norm.}
The center-vector norm is computed after row-wise $L_2$ normalization~\cite{jha2024commonstabilitymechanismselfsupervised}:
\[
    \operatorname{CVN}(X)=
    \left\lVert\frac{1}{N}\sum_{i=1}^{N}
    \frac{x_i}{\lVert x_i\rVert_2}\right\rVert_2.
\]

\paragraph{Feature correlation.}
Feature correlation is the mean absolute upper-triangular off-diagonal entry of the Pearson correlation matrix $R$ of the columns of $X$. Undefined correlations from zero-variance dimensions are replaced by zero. For $\mathcal{U}=\{(p,q):p<q\}$,
\[
    \operatorname{Corr}(X)=
    \frac{1}{|\mathcal{U}|}\sum_{(p,q)\in\mathcal{U}}|R_{pq}|.
\]

\paragraph{Effective rank.}
For the matrix being evaluated, let $\sigma_1,\ldots,\sigma_r$ be its singular values and define
\[
    p_k=\frac{\sigma_k}{\sum_l\sigma_l}.
\]
We use the entropy-based effective rank~\cite{roy2007rank}:
\[
    \operatorname{erank}(X)=
    \exp\left(-\sum_{k=1}^{r}p_k\log p_k\right).
\]
In our implementation, effective rank is computed on $X$ without centering, and singular-value probabilities smaller than $10^{-10}$ are discarded before evaluating the entropy. We compute this quantity separately for the matrix of class centroids and for the full embedding matrix.

\paragraph{Indicator function.}
For any condition $A$, the indicator function is defined as
\[
    \mathbf{1}[A]=
    \begin{cases}
        1, & \text{if } A \text{ is true},\\
        0, & \text{otherwise}.
    \end{cases}
\]

\paragraph{Sparsity.}
Sparsity is the fraction of entries whose absolute activation is below $\tau=10^{-4}$:
\[
    \operatorname{Sparsity}(X)=
    \frac{1}{ND}\sum_{i=1}^{N}\sum_{k=1}^{D}
    \mathbf{1}[|X_{ik}|<\tau].
\]

\paragraph{Dimension selectivity.}
Let $C$ be the number of classes and let $\mathcal{I}_c$ denote the set of samples belonging to class $c$, for $c=1,\ldots,C$. We first compute the class centroids:
\[
    m_{ck}=|\mathcal{I}_c|^{-1}\sum_{i\in\mathcal{I}_c}X_{ik}.
\]
For each dimension $k$, we compute the mean $\mu_k$ and population standard deviation $s_k$ of the centroid values $\{m_{ck}\}_{c=1}^{C}$, and standardize each dimension over classes:
\[
    z_{ck}=\frac{m_{ck}-\mu_k}{s_k+10^{-8}}.
\]
A dimension $k$ is counted as active for class $c$ when $z_{ck}>3$. The reported bins count dimensions according to
\[
    n_k=\sum_{c=1}^{C}\mathbf{1}[z_{ck}>3].
\]

\paragraph{Sensitivity index.}
For positive-pair similarities $S^+$ and negative-pair similarities $S^-$, the sensitivity index is
\[
    d^\prime=\frac{\mu_+-\mu_-}
    {\sqrt{\frac{\sigma_+^2+\sigma_-^2}{2}}},
\]
where $\mu_\pm$ and $\sigma_\pm^2$ are the corresponding means and population variances. All reported pairwise quantities exclude self-pairs.

\subsubsection{Global Properties.}
Using our networks trained on ImageNet-1k and Food-101, we measured several quantitative latent space properties, summarized in \cref{tab:qualityMetrics}. Surprisingly, even though our method relies only on local repulsion, Hypersolid learned a geometry with globally low anisotropy, low dimensional correlation, and low center vector norm~\cite{jha2024commonstabilitymechanismselfsupervised}, suggesting it has learned well-distributed representations across the latent space. This contrasts with methods such as Barlow Twins or VICReg, which explicitly optimize the global distribution of representations.

\begin{table}[htbp]
\caption{Percent of dimension selectivity on ImageNet-1k (ResNet-50). Features are grouped by the number of classes they activate.}
\label{tab:dimension_selectivity}
\centering
\small
\setlength{\tabcolsep}{8pt}
\begin{tabular}{@{}lrrrr@{}}
\toprule
\bfseries Method & \bfseries 0 & \bfseries 1-5 & \bfseries 6-20 & \bfseries 21+ \\
\midrule
Supervised      &  0.0 &  0.0 & 70.5 & 29.5 \\
SimCLR          &  0.2 &  3.1 & 67.5 & 29.2 \\
BYOL            &  0.0 & 14.6 & 54.7 & 30.7 \\
SwAV            &  0.0 &  1.3 & 60.8 & 37.9 \\
Barlow Twins    &  0.3 &  4.6 & 66.2 & 28.9 \\
DINO            &  0.1 &  3.6 & 78.6 & 17.6 \\
VICReg          &  0.3 &  2.5 & 54.5 & 42.6 \\
ReSA            &  0.1 &  2.6 & 67.7 & 29.6 \\
Hypersolid      &  0.3 &  8.5 & 73.7 & 17.4 \\
--- w/o L2      &  0.0 &  0.4 & 44.4 & 55.1 \\
\bottomrule
\end{tabular}
\end{table}

A particularity of Hypersolid is that it has comparatively lower effective rank both for class centroids and for embeddings, together with much higher sparsity. This suggests that Hypersolid is not necessarily learning a more descriptive representation, but rather it is clustering similar samples into highly sparse representations. This is partially confirmed by the dimensional selectivity analysis in \cref{tab:dimension_selectivity}. Compared with other methods, Hypersolid has a comparatively higher number of features that are activated by a small number of classes, and a lower number of features that are activated by many classes. This is consistent with being better suited for fine-grained classification, but also suggests that Hypersolid is less suited for general-purpose transfer learning. On the other hand, the version of Hypersolid without L2 regularization has a comparatively higher number of features that are activated by many classes, suggesting that this version may be more suited for general-purpose transfer learning, at the cost of having a less sparse representation and less tight clusters.

\subsubsection{Pairwise Similarities.}

Complementing the global metrics, we analyze the distribution of pairwise cosine similarities, which directly informs how well different classes can be distinguished in the latent space. To quantify pairwise separability, we employ the Sensitivity Index (\(d^\prime\)) from Signal Detection Theory~\cite{Macmillan2004}. The sensitivity index measures the statistical separation between the distribution of positive and negative pair similarities, normalized by their pooled variance. As shown in \Cref{tab:qualityMetrics}, in these comparisons Hypersolid achieved a higher sensitivity index than other SSL methods and even the supervised baseline.

\begin{figure}[htbp]
\centering
\begin{tikzpicture}
    \pgfplotsset{
        kde pos/.style={smooth, thick, draw=blue, fill=blue!20, fill opacity=0.5},
        kde neg/.style={smooth, thick, draw=orange, fill=orange!20, fill opacity=0.5},
    }

    \pgfplotstableread[col sep=comma]{data/kde_pairwise_similarities_im1k.txt}\data
    \begin{groupplot}[
        group style={
            group size=2 by 5,
            vertical sep=0.2cm,
            horizontal sep=0.2cm,
            xlabels at=edge bottom,
            xticklabels at=edge bottom,
            ylabels at=edge left,
            yticklabels at=edge left,
        },
        width=0.55\textwidth,
        height=3.5cm,
        xmin=-0.05, xmax=1,
        ymin=0, ymax=7.5,
        ytick={2.5, 5, 7.5},
        xlabel={Cosine Similarity},
        grid style={dashed, gray!20},
        ymajorgrids=true,
        xmajorgrids=true,
        grid=major,
        axis line style={draw=none},
        tick style={draw=none},
        label style={font=\scriptsize},
        tick label style={font=\scriptsize},
        legend style={draw=none, fill=none, font=\scriptsize, /tikz/every even column/.append style={column sep=0.5cm}},
        legend columns=2,
    ]

    \nextgroupplot
    \node[anchor=north] at (axis description cs:0.5,0.95) {\small Supervised};
    \addplot[kde pos] table[col sep=comma, x=similarity, y=positive_supervised_resnet50] {\data} \closedcycle;
    \addplot[kde neg] table[col sep=comma, x=similarity, y=negative_supervised_resnet50] {\data} \closedcycle;

    \nextgroupplot
    \node[anchor=north] at (axis description cs:0.5,0.95) {\small SimCLR};
    \addplot[kde pos] table[col sep=comma, x=similarity, y=positive_simclr_resnet50] {\data} \closedcycle;
    \addplot[kde neg] table[col sep=comma, x=similarity, y=negative_simclr_resnet50] {\data} \closedcycle;

    \nextgroupplot
    \node[anchor=north] at (axis description cs:0.5,0.95) {\small BYOL};
    \addplot[kde pos] table[col sep=comma, x=similarity, y=positive_byol_resnet50] {\data} \closedcycle;
    \addplot[kde neg] table[col sep=comma, x=similarity, y=negative_byol_resnet50] {\data} \closedcycle;

    \nextgroupplot
    \node[anchor=north] at (axis description cs:0.5,0.95) {\small SwAV};
    \addplot[kde pos] table[col sep=comma, x=similarity, y=positive_swav_resnet50] {\data} \closedcycle;
    \addplot[kde neg] table[col sep=comma, x=similarity, y=negative_swav_resnet50] {\data} \closedcycle;

    \nextgroupplot
    \node[anchor=north] at (axis description cs:0.5,0.95) {\small Barlow Twins};
    \addplot[kde pos] table[col sep=comma, x=similarity, y=positive_barlow_twins_resnet50] {\data} \closedcycle;
    \addplot[kde neg] table[col sep=comma, x=similarity, y=negative_barlow_twins_resnet50] {\data} \closedcycle;

    \nextgroupplot
    \node[anchor=north] at (axis description cs:0.5,0.95) {\small VICReg};
    \addplot[kde pos] table[col sep=comma, x=similarity, y=positive_vicreg_resnet50] {\data} \closedcycle;
    \addplot[kde neg] table[col sep=comma, x=similarity, y=negative_vicreg_resnet50] {\data} \closedcycle;

    \nextgroupplot
    \node[anchor=north] at (axis description cs:0.35,0.95) {\small DINO};
    \addplot[kde pos] table[col sep=comma, x=similarity, y=positive_dino_resnet50] {\data} \closedcycle;
    \addplot[kde neg] table[col sep=comma, x=similarity, y=negative_dino_resnet50] {\data} \closedcycle;

    \nextgroupplot
    \node[anchor=north] at (axis description cs:0.5,0.95) {\small ReSA};
    \addplot[kde pos] table[col sep=comma, x=similarity, y=positive_resa_resnet50] {\data} \closedcycle;
    \addplot[kde neg] table[col sep=comma, x=similarity, y=negative_resa_resnet50] {\data} \closedcycle;

    \nextgroupplot
    \node[anchor=north] at (axis description cs:0.5,0.95) {\small Hypersolid};
    \addplot[kde pos] table[col sep=comma, x=similarity, y=positive_hypersolid_maxpool_norm_resnet50] {\data} \closedcycle;
    \addplot[kde neg] table[col sep=comma, x=similarity, y=negative_hypersolid_maxpool_norm_resnet50] {\data} \closedcycle;

    \nextgroupplot
    \node[anchor=north] at (axis description cs:0.5,0.95) {\small Hypersolid w/o L2};
    \addplot[kde pos] table[col sep=comma, x=similarity, y=positive_hypersolid_maxpool_resnet50] {\data} \closedcycle;
    \addplot[kde neg] table[col sep=comma, x=similarity, y=negative_hypersolid_maxpool_resnet50] {\data} \closedcycle;

    \end{groupplot}

    \node[rotate=90, anchor=center] at ($(group c1r1.north west)!0.5!(group c1r5.south west) + (-0.8cm, 0)$) {\small Density};

    \node[anchor=north, font=\scriptsize] at ($(group c1r5.south)!0.5!(group c2r5.south) + (0,-0.7cm)$) {%
        \setlength{\fboxsep}{1.2pt}%
        \fcolorbox{blue}{blue!20}{\rule{0pt}{1.05ex}\hspace{0.7em}}\,\textcolor{black}{Positive Pairs}%
        \hspace{1.4em}%
        \fcolorbox{orange}{orange!20}{\rule{0pt}{1.05ex}\hspace{0.7em}}\,\textcolor{black}{Negative Pairs}%
    };
\end{tikzpicture}
\caption{Density of Pairwise Cosine Similarity (ImageNet-1k).}
\label{fig:density_stack}
\end{figure}

\begin{figure*}[!htbp]
    \centering

    \newlength{\panelw}
    \setlength{\panelw}{0.18\textwidth}

    \subfloat[Supervised]{
        \includegraphics[width=\panelw]{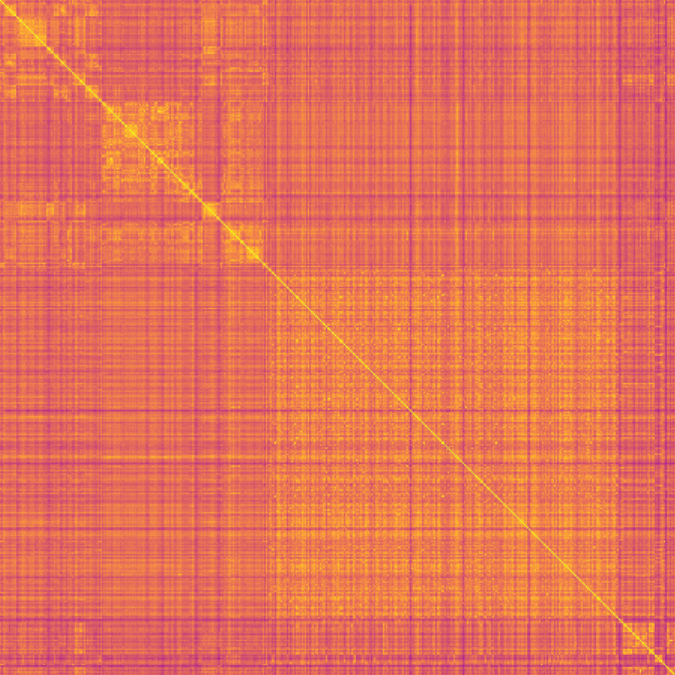}
    }\hfill
    \subfloat[SimCLR]{
        \includegraphics[width=\panelw]{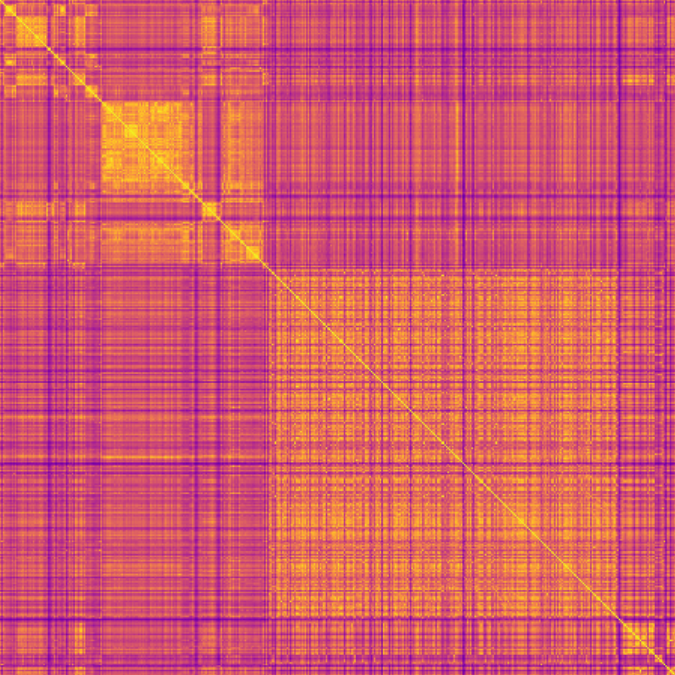}
    }\hfill
    \subfloat[BYOL]{
        \includegraphics[width=\panelw]{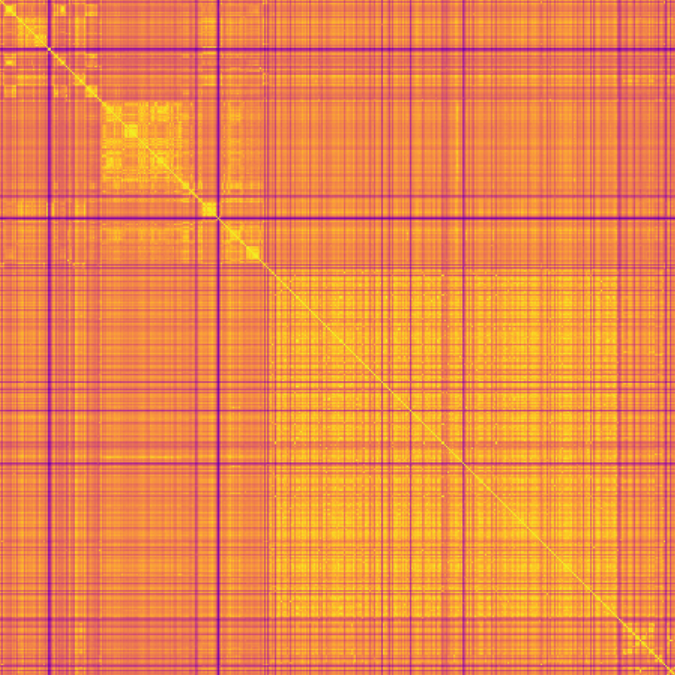}
    }\hfill
    \subfloat[SwAV]{
        \includegraphics[width=\panelw]{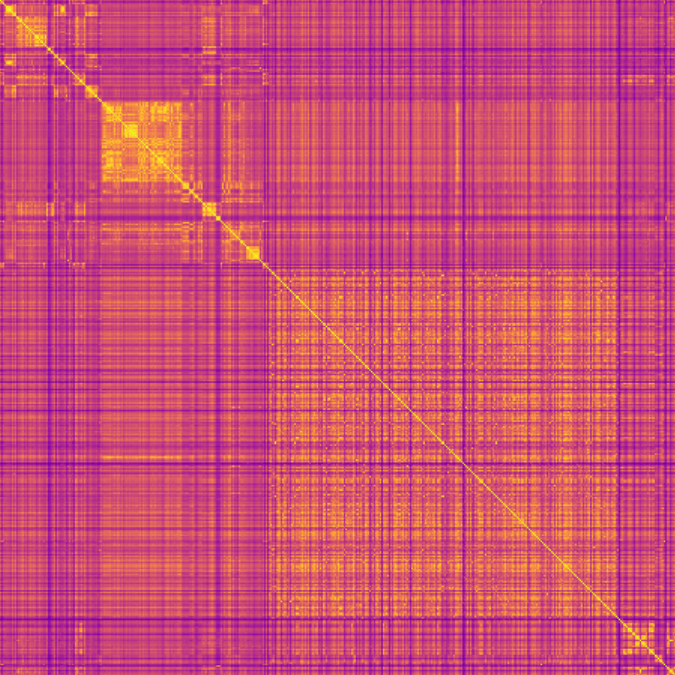}
    }\hfill
    \subfloat[Barlow Twins]{
        \includegraphics[width=\panelw]{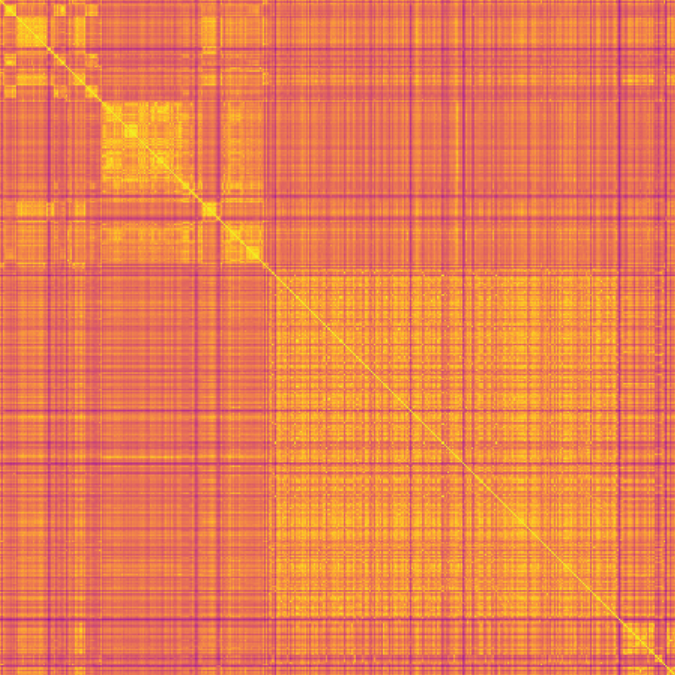}
    }

    \vspace{0.6em}

    \makebox[\textwidth][c]{%
        \subfloat[DINO]{
            \includegraphics[width=\panelw]{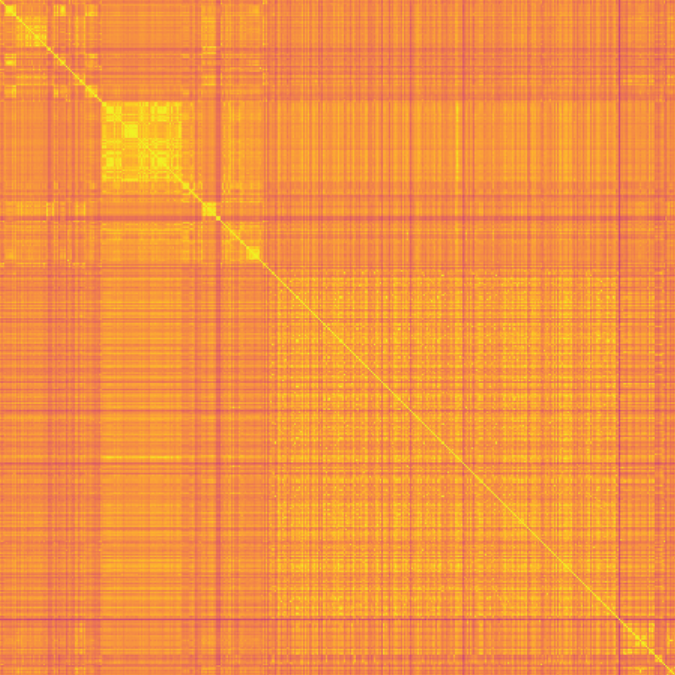}
        }\hspace{0.005\textwidth}
        \subfloat[VICReg]{
            \includegraphics[width=\panelw]{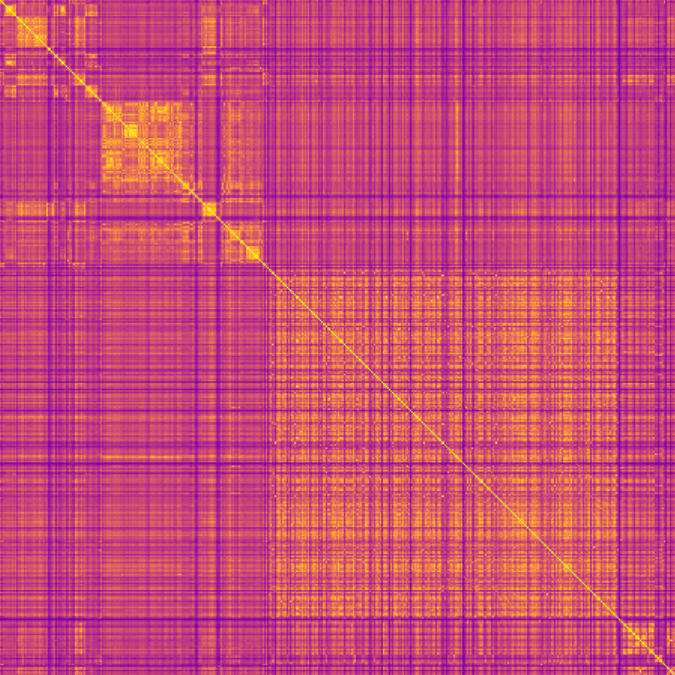}
        }\hspace{0.005\textwidth}
        \subfloat[ReSA]{
            \includegraphics[width=\panelw]{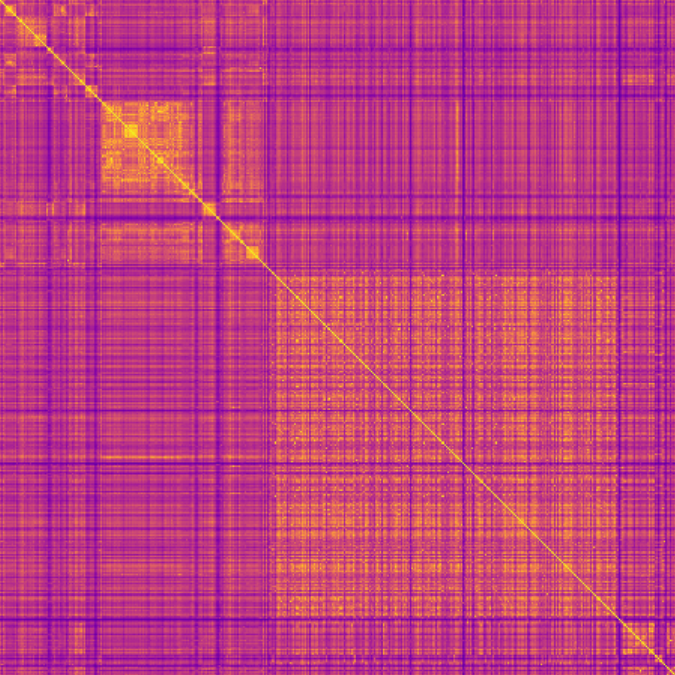}
        }\hspace{0.005\textwidth}
        \subfloat[Hypersolid]{
            \includegraphics[width=\panelw]{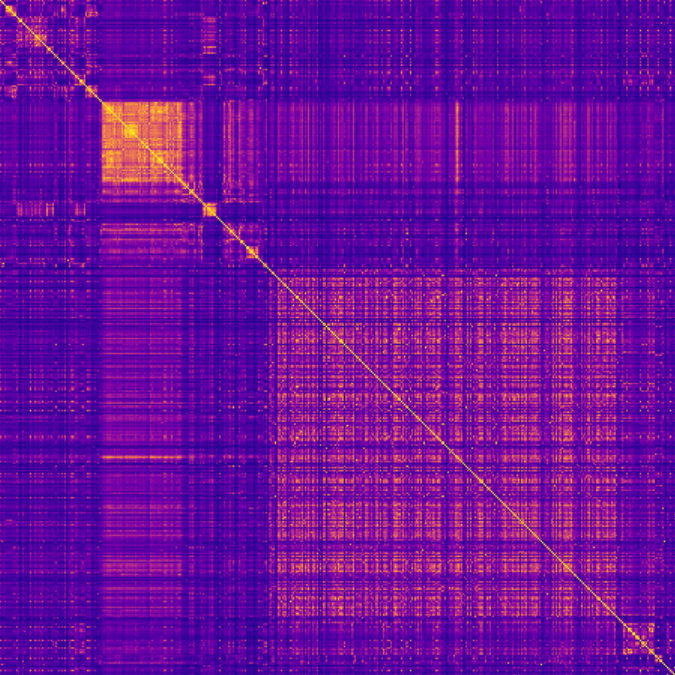}
        }\hspace{0.04\textwidth}
        \makebox[0.16\panelw][c]{%
            \raisebox{0.42\panelw}[0pt][0pt]{%
                \plasmacolorbar{6}{70}%
            }%
        }
    }

    \caption{Pairwise cosine similarity between ImageNet-1k class centroids. Brighter means more similarity, while darker colors mean bigger angles (0 is equivalent to $90^\circ$).}
    \label{fig:centroid_pairwise_similarity_all}
\end{figure*}

The geometric intuition behind sensitivity can be observed in \Cref{fig:density_stack}, which presents a comprehensive comparison of pairwise cosine similarities across all evaluated methods, separated by positive and negative pairs. For instance, SimCLR, SwAV, VICReg, and ReSA show a similar distribution of positive pairs, with density increasing following a ``bell-shaped'' curve, while the negatives are concentrated in a region with low cosine similarity but still far from orthogonality. In contrast, Hypersolid, both with and without L2 regularization, displays a more uniform distribution of positive pairs, while the negative pairs are concentrated in a region with low cosine similarity. Interestingly, the L2 regularization moved the density of negative pairs closer to orthogonality. This figure helps to visualize geometrically how differences in sensitivity index ($d^\prime$) can impact class distinguishability.

\Cref{fig:centroid_pairwise_similarity_all} shows the pairwise cosine similarity between class centroids for all methods. The heatmaps show a clear difference between Hypersolid and other methods: Hypersolid centroid similarity with other classes is concentrated closer to zero (closer to orthogonality), while other methods such as DINO or BYOL show a much higher degree of similarity between centroids of different classes. This suggests that features learned by Hypersolid are more class-specific, which would also explain the strong clustering performance reported in \cref{subsec:clustering_properties}. In contrast, other methods' heatmaps show a higher degree of similarity between different classes, which coincides with higher anisotropy. Recent work indicates that anisotropy can help feature reuse \cite{luthra2026directionalneuralcollapseexplains}, so our results align with this finding, as Hypersolid geometry dimensions seem to be class aligned, without strong feature sharing across classes.

Both Hypersolid networks were trained using $\alpha=0.9$, meaning the optimization enforced only a minimal angular distance of around $25.84^\circ$ between pairs. Surprisingly, the resulting geometry for Hypersolid ended up with much larger mean pairwise angles than other methods, with $83.47^\circ$ on ImageNet-1k and $76.51^\circ$ on Food-101. This signals that, even if it may be counterintuitive, when the goal is to maximize distances between all pairs, it is a better policy to enforce only a minimal distance and let the points self-organize.

\subsubsection{Neighborhood Properties.} To characterize the local geometry of Hypersolid, we performed linear interpolations between embeddings of randomly selected ImageNet-1k image pairs, and measured the cosine distance from each interpolated point to the nearest validation-set embedding. As illustrated in \cref{fig:energy_2x2}, Hypersolid exhibits greater visible separation between inter-class (negative) and intra-class (positive) interpolation paths. This suggests the existence of low-density gaps between class neighborhoods, which may facilitate cluster discrimination, as will be analyzed next.

\begin{figure}[htbp]
\centering
\begin{tikzpicture}
    \pgfplotsset{
        energy boundary/.style={draw=none, forget plot},
        energy neg fill/.style={orange!35, opacity=0.8, forget plot},
        energy neg line/.style={thick, orange},
        energy pos fill/.style={blue!35, opacity=0.8, forget plot},
        energy pos line/.style={thick, blue},
    }

    \pgfplotstableread[col sep=comma]{data/pairwise_energy_profiles.txt}\data
    \begin{groupplot}[
        group style={
            group size=2 by 5,
            horizontal sep=0.4cm,
            vertical sep=0.5cm,
            xlabels at=edge bottom,
            ylabels at=edge left,
            yticklabels at=edge left,
        },
        width=0.55\textwidth,
        height=3cm,
        xmin=0, xmax=1,
        ymin=0, ymax=0.30,
        grid=major,
        grid style={dashed, gray!20},
        axis line style={draw=none},
        tick style={draw=none},
        label style={font=\scriptsize},
        tick label style={font=\scriptsize},
        title style={font=\footnotesize, yshift=-0.3cm},
        xtick={0, 0.5, 1},
        xticklabels={A, $t$, B},
        legend style={draw=none, fill=none, font=\scriptsize, /tikz/every even column/.append style={column sep=0.5cm}},
        legend columns=2,
    ]

    \nextgroupplot
    \node[anchor=north west] at (axis description cs:0.0,0.95) {\small Supervised};
    \addplot [name path=nr, energy boundary] table [x=supervised_resnet50_step, y=supervised_resnet50_neg_upper] {\data};
    \addplot [name path=nr2, energy boundary] table [x=supervised_resnet50_step, y=supervised_resnet50_neg_lower] {\data};
    \addplot [energy neg fill] fill between [of=nr and nr2];
    \addplot [energy neg line] table [x=supervised_resnet50_step, y=supervised_resnet50_neg_mean] {\data};
    \addplot [name path=pr, energy boundary] table [x=supervised_resnet50_step, y=supervised_resnet50_pos_upper] {\data};
    \addplot [name path=pr2, energy boundary] table [x=supervised_resnet50_step, y=supervised_resnet50_pos_lower] {\data};
    \addplot [energy pos fill] fill between [of=pr and pr2];
    \addplot [energy pos line] table [x=supervised_resnet50_step, y=supervised_resnet50_pos_mean] {\data};

    \nextgroupplot
    \node[anchor=north west] at (axis description cs:0.0,0.95) {\small SimCLR};
    \addplot [name path=nr, energy boundary] table [x=simclr_resnet50_step, y=simclr_resnet50_neg_upper] {\data};
    \addplot [name path=nr2, energy boundary] table [x=simclr_resnet50_step, y=simclr_resnet50_neg_lower] {\data};
    \addplot [energy neg fill] fill between [of=nr and nr2];
    \addplot [energy neg line] table [x=simclr_resnet50_step, y=simclr_resnet50_neg_mean] {\data};
    \addplot [name path=pr, energy boundary] table [x=simclr_resnet50_step, y=simclr_resnet50_pos_upper] {\data};
    \addplot [name path=pr2, energy boundary] table [x=simclr_resnet50_step, y=simclr_resnet50_pos_lower] {\data};
    \addplot [energy pos fill] fill between [of=pr and pr2];
    \addplot [energy pos line] table [x=simclr_resnet50_step, y=simclr_resnet50_pos_mean] {\data};

    \nextgroupplot
    \node[anchor=north west] at (axis description cs:0.0,0.95) {\small BYOL};
    \addplot [name path=nr, energy boundary] table [x=byol_resnet50_step, y=byol_resnet50_neg_upper] {\data};
    \addplot [name path=nr2, energy boundary] table [x=byol_resnet50_step, y=byol_resnet50_neg_lower] {\data};
    \addplot [energy neg fill] fill between [of=nr and nr2];
    \addplot [energy neg line] table [x=byol_resnet50_step, y=byol_resnet50_neg_mean] {\data};
    \addplot [name path=pr, energy boundary] table [x=byol_resnet50_step, y=byol_resnet50_pos_upper] {\data};
    \addplot [name path=pr2, energy boundary] table [x=byol_resnet50_step, y=byol_resnet50_pos_lower] {\data};
    \addplot [energy pos fill] fill between [of=pr and pr2];
    \addplot [energy pos line] table [x=byol_resnet50_step, y=byol_resnet50_pos_mean] {\data};

    \nextgroupplot
    \node[anchor=north west] at (axis description cs:0.0,0.95) {\small SwAV};
    \addplot [name path=nr, energy boundary] table [x=swav_resnet50_step, y=swav_resnet50_neg_upper] {\data};
    \addplot [name path=nr2, energy boundary] table [x=swav_resnet50_step, y=swav_resnet50_neg_lower] {\data};
    \addplot [energy neg fill] fill between [of=nr and nr2];
    \addplot [energy neg line] table [x=swav_resnet50_step, y=swav_resnet50_neg_mean] {\data};
    \addplot [name path=pr, energy boundary] table [x=swav_resnet50_step, y=swav_resnet50_pos_upper] {\data};
    \addplot [name path=pr2, energy boundary] table [x=swav_resnet50_step, y=swav_resnet50_pos_lower] {\data};
    \addplot [energy pos fill] fill between [of=pr and pr2];
    \addplot [energy pos line] table [x=swav_resnet50_step, y=swav_resnet50_pos_mean] {\data};

    \nextgroupplot
    \node[anchor=north west] at (axis description cs:0.0,0.95) {\small Barlow Twins};
    \addplot [name path=nr, energy boundary] table [x=barlow_twins_resnet50_step, y=barlow_twins_resnet50_neg_upper] {\data};
    \addplot [name path=nr2, energy boundary] table [x=barlow_twins_resnet50_step, y=barlow_twins_resnet50_neg_lower] {\data};
    \addplot [energy neg fill] fill between [of=nr and nr2];
    \addplot [energy neg line] table [x=barlow_twins_resnet50_step, y=barlow_twins_resnet50_neg_mean] {\data};
    \addplot [name path=pr, energy boundary] table [x=barlow_twins_resnet50_step, y=barlow_twins_resnet50_pos_upper] {\data};
    \addplot [name path=pr2, energy boundary] table [x=barlow_twins_resnet50_step, y=barlow_twins_resnet50_pos_lower] {\data};
    \addplot [energy pos fill] fill between [of=pr and pr2];
    \addplot [energy pos line] table [x=barlow_twins_resnet50_step, y=barlow_twins_resnet50_pos_mean] {\data};

    \nextgroupplot
    \node[anchor=north west] at (axis description cs:0.0,0.95) {\small VICReg};
    \addplot [name path=nr, energy boundary] table [x=vicreg_resnet50_step, y=vicreg_resnet50_neg_upper] {\data};
    \addplot [name path=nr2, energy boundary] table [x=vicreg_resnet50_step, y=vicreg_resnet50_neg_lower] {\data};
    \addplot [energy neg fill] fill between [of=nr and nr2];
    \addplot [energy neg line] table [x=vicreg_resnet50_step, y=vicreg_resnet50_neg_mean] {\data};
    \addplot [name path=pr, energy boundary] table [x=vicreg_resnet50_step, y=vicreg_resnet50_pos_upper] {\data};
    \addplot [name path=pr2, energy boundary] table [x=vicreg_resnet50_step, y=vicreg_resnet50_pos_lower] {\data};
    \addplot [energy pos fill] fill between [of=pr and pr2];
    \addplot [energy pos line] table [x=vicreg_resnet50_step, y=vicreg_resnet50_pos_mean] {\data};

    \nextgroupplot
    \node[anchor=north west] at (axis description cs:0.0,0.95) {\small DINO};
    \addplot [name path=nr, energy boundary] table [x=dino_resnet50_step, y=dino_resnet50_neg_upper] {\data};
    \addplot [name path=nr2, energy boundary] table [x=dino_resnet50_step, y=dino_resnet50_neg_lower] {\data};
    \addplot [energy neg fill] fill between [of=nr and nr2];
    \addplot [energy neg line] table [x=dino_resnet50_step, y=dino_resnet50_neg_mean] {\data};
    \addplot [name path=pr, energy boundary] table [x=dino_resnet50_step, y=dino_resnet50_pos_upper] {\data};
    \addplot [name path=pr2, energy boundary] table [x=dino_resnet50_step, y=dino_resnet50_pos_lower] {\data};
    \addplot [energy pos fill] fill between [of=pr and pr2];
    \addplot [energy pos line] table [x=dino_resnet50_step, y=dino_resnet50_pos_mean] {\data};

    \nextgroupplot
    \node[anchor=north west] at (axis description cs:0.0,0.95) {\small ReSA};
    \addplot [name path=nr, energy boundary] table [x=resa_resnet50_step, y=resa_resnet50_neg_upper] {\data};
    \addplot [name path=nr2, energy boundary] table [x=resa_resnet50_step, y=resa_resnet50_neg_lower] {\data};
    \addplot [energy neg fill] fill between [of=nr and nr2];
    \addplot [energy neg line] table [x=resa_resnet50_step, y=resa_resnet50_neg_mean] {\data};
    \addplot [name path=pr, energy boundary] table [x=resa_resnet50_step, y=resa_resnet50_pos_upper] {\data};
    \addplot [name path=pr2, energy boundary] table [x=resa_resnet50_step, y=resa_resnet50_pos_lower] {\data};
    \addplot [energy pos fill] fill between [of=pr and pr2];
    \addplot [energy pos line] table [x=resa_resnet50_step, y=resa_resnet50_pos_mean] {\data};

    \nextgroupplot
    \node[anchor=north west] at (axis description cs:0.0,0.95) {\small Hypersolid};
    \addplot [name path=nr, energy boundary] table [x=hypersolid_maxpool_norm_resnet50_step, y=hypersolid_maxpool_norm_resnet50_neg_upper] {\data};
    \addplot [name path=nr2, energy boundary] table [x=hypersolid_maxpool_norm_resnet50_step, y=hypersolid_maxpool_norm_resnet50_neg_lower] {\data};
    \addplot [energy neg fill] fill between [of=nr and nr2];
    \addplot [energy neg line] table [x=hypersolid_maxpool_norm_resnet50_step, y=hypersolid_maxpool_norm_resnet50_neg_mean] {\data};
    \addplot [name path=pr, energy boundary] table [x=hypersolid_maxpool_norm_resnet50_step, y=hypersolid_maxpool_norm_resnet50_pos_upper] {\data};
    \addplot [name path=pr2, energy boundary] table [x=hypersolid_maxpool_norm_resnet50_step, y=hypersolid_maxpool_norm_resnet50_pos_lower] {\data};
    \addplot [energy pos fill] fill between [of=pr and pr2];
    \addplot [energy pos line] table [x=hypersolid_maxpool_norm_resnet50_step, y=hypersolid_maxpool_norm_resnet50_pos_mean] {\data};

    \nextgroupplot
    \node[anchor=north west] at (axis description cs:0.0,0.95) {\small w/o L2};
    \addplot [name path=nr, energy boundary] table [x=hypersolid_maxpool_resnet50_step, y=hypersolid_maxpool_resnet50_neg_upper] {\data};
    \addplot [name path=nr2, energy boundary] table [x=hypersolid_maxpool_resnet50_step, y=hypersolid_maxpool_resnet50_neg_lower] {\data};
    \addplot [energy neg fill] fill between [of=nr and nr2];
    \addplot [energy neg line] table [x=hypersolid_maxpool_resnet50_step, y=hypersolid_maxpool_resnet50_neg_mean] {\data};
    \addplot [name path=pr, energy boundary] table [x=hypersolid_maxpool_resnet50_step, y=hypersolid_maxpool_resnet50_pos_upper] {\data};
    \addplot [name path=pr2, energy boundary] table [x=hypersolid_maxpool_resnet50_step, y=hypersolid_maxpool_resnet50_pos_lower] {\data};
    \addplot [energy pos fill] fill between [of=pr and pr2];
    \addplot [energy pos line] table [x=hypersolid_maxpool_resnet50_step, y=hypersolid_maxpool_resnet50_pos_mean] {\data};

    \end{groupplot}

    \node[rotate=90, anchor=center] at ($(group c1r1.north west)!0.5!(group c1r5.south west) + (-0.8cm, 0)$) {\small Cosine Distance to Nearest Neighbor};

    \node[anchor=north, font=\scriptsize] at ($(group c1r5.south)!0.5!(group c2r5.south) + (0,-0.7cm)$) {%
        \setlength{\fboxsep}{1.2pt}%
        \fcolorbox{orange}{orange!35}{\rule{0pt}{1.05ex}\hspace{0.7em}}\,\textcolor{black}{Negative Pairs}%
        \hspace{1.4em}%
        \fcolorbox{blue}{blue!35}{\rule{0pt}{1.05ex}\hspace{0.7em}}\,\textcolor{black}{Positive Pairs}%
    };
\end{tikzpicture}
\caption{Nearest-neighbor cosine distance along interpolations between random embedding pairs. Solid lines show the mean distance; shaded regions indicate $\pm 1$ standard deviation.}
\label{fig:energy_2x2}
\end{figure}

\subsection{Clustering}\label{subsec:clustering_properties}
Our findings regarding pairwise separability and neighborhood properties suggest that Hypersolid may be particularly well-suited for clustering tasks. To further investigate this, we conducted a series of experiments using K-Means, DBSCAN \cite{ester1996dbscan} and HDBSCAN \cite{campello2013hdbscan}. For DBSCAN, we explored the impact of varying the EPS parameter on clustering performance, as measured by Normalized Mutual Information (NMI) and Adjusted Rand Index (ARI). The results are presented in \cref{fig:DBSCAN_NMI} and \cref{fig:DBSCAN_ARI}, respectively. For K-Means and HDBSCAN, we measured NMI, ARI and the number of clusters found (for HDBSCAN), with results summarized in \cref{tab:clustering-results}.

As shown in \cref{fig:DBSCAN_NMI}, Hypersolid (both with and without L2 regularization) exhibits much higher NMI at low EPS values compared to other methods, indicating that Hypersolid clusters are more tightly packed and well-separated. Other methods reach their peak NMI at higher EPS values, but still do not reach the level of Hypersolid. This suggests that clustering Hypersolid embeddings results in clusters better aligned with the ground truth classes, at least on ImageNet-1k. This is consistent with our ARI measurements in \cref{fig:DBSCAN_ARI}, where Hypersolid also directionally outperforms other methods across a range of EPS values. In this case, the absolute ARI values are low; however, we must consider that we are evaluating clusters against a ground truth with 1000 classes using a totally unsupervised approach. The best NMI and ARI values over the EPS sweep, together with their corresponding EPS values, are reported in \cref{tab:dbscan-best}.

\begin{figure}[!htbp]
\centering
\begin{tikzpicture}
\begin{axis}[
    ymin=0.0,
    xmin=0,
    xmax=0.5,
    width=0.9\columnwidth,
    height=0.6\columnwidth,
    xlabel={EPS},
    ylabel={NMI},
    xtick={0,0.1,0.2,0.3,0.4,0.5},
    xticklabel style={
        /pgf/number format/fixed,
        /pgf/number format/precision=1
    },
    grid=major,
    legend columns=5,
    legend style={
        at={(0.5,-0.2)},
        anchor=north,
        draw=none,
        font=\footnotesize
    },
    cycle list name=mycolorlist,
    every axis plot post/.append style={
        mark repeat=20,
        mark phase=1
    }
]

\addplot table [x=EPS, y=Supervised_NMI] {data/dbscan_eps_exploration.txt};
\addlegendentry{Supervised}

\addplot table [x=EPS, y=SimCLR_NMI] {data/dbscan_eps_exploration.txt};
\addlegendentry{SimCLR}

\addplot table [x=EPS, y=BYOL_NMI] {data/dbscan_eps_exploration.txt};
\addlegendentry{BYOL}

\addplot table [x=EPS, y=SwAV_NMI] {data/dbscan_eps_exploration.txt};
\addlegendentry{SwAV}

\addplot table [x=EPS, y=Barlow_Twins_NMI] {data/dbscan_eps_exploration.txt};
\addlegendentry{Barlow Twins}

\addplot table [x=EPS, y=VICReg_NMI] {data/dbscan_eps_exploration.txt};
\addlegendentry{VICReg}

\addplot table [x=EPS, y=DINO_NMI] {data/dbscan_eps_exploration.txt};
\addlegendentry{DINO}


\addplot table [x=EPS, y=ReSA_NMI] {data/dbscan_eps_exploration.txt};
\addlegendentry{ReSA}

\addplot table [x=EPS, y=Hypersolid_MaxPool_Norm_NMI] {data/dbscan_eps_exploration.txt};
\addlegendentry{Hypersolid}

\addplot table [x=EPS, y=Hypersolid_MaxPool_NMI] {data/dbscan_eps_exploration.txt};
\addlegendentry{Hypersolid w/o L2}

\end{axis}
\end{tikzpicture}
\caption{DBSCAN with cosine metric NMI as a function of EPS.}
\label{fig:DBSCAN_NMI}
\end{figure}

\begin{figure}[!htbp]
\centering
\begin{tikzpicture}
\begin{axis}[
    ymin=0.0,
    xmin=0,
    xmax=0.5,
    width=0.9\columnwidth,
    height=0.6\columnwidth,
    xlabel={EPS},
    ylabel={ARI},
    xtick={0,0.1,0.2,0.3,0.4,0.5},
    xticklabel style={
        /pgf/number format/fixed,
        /pgf/number format/precision=1
    },
    grid=major,
    legend columns=5,
    legend style={
        at={(0.5,-0.2)},
        anchor=north,
        draw=none,
        font=\footnotesize
    },
    cycle list name=mycolorlist,
    every axis plot post/.append style={
        mark repeat=20,
        mark phase=1
    }
]

\addplot table [x=EPS, y=Supervised_ARI] {data/dbscan_eps_exploration.txt};
\addlegendentry{Supervised}

\addplot table [x=EPS, y=SimCLR_ARI] {data/dbscan_eps_exploration.txt};
\addlegendentry{SimCLR}

\addplot table [x=EPS, y=BYOL_ARI] {data/dbscan_eps_exploration.txt};
\addlegendentry{BYOL}

\addplot table [x=EPS, y=SwAV_ARI] {data/dbscan_eps_exploration.txt};
\addlegendentry{SwAV}

\addplot table [x=EPS, y=Barlow_Twins_ARI] {data/dbscan_eps_exploration.txt};
\addlegendentry{Barlow Twins}

\addplot table [x=EPS, y=VICReg_ARI] {data/dbscan_eps_exploration.txt};
\addlegendentry{VICReg}

\addplot table [x=EPS, y=DINO_ARI] {data/dbscan_eps_exploration.txt};
\addlegendentry{DINO}


\addplot table [x=EPS, y=ReSA_ARI] {data/dbscan_eps_exploration.txt};
\addlegendentry{ReSA}

\addplot table [x=EPS, y=Hypersolid_MaxPool_Norm_ARI] {data/dbscan_eps_exploration.txt};
\addlegendentry{Hypersolid}

\addplot table [x=EPS, y=Hypersolid_MaxPool_ARI] {data/dbscan_eps_exploration.txt};
\addlegendentry{Hypersolid w/o L2}

\end{axis}
\end{tikzpicture}
\caption{DBSCAN with cosine metric ARI as a function of EPS.}
\label{fig:DBSCAN_ARI}
\end{figure}

\begin{table}[!htbp]
\caption{Best DBSCAN clustering NMI-EPS and ARI-EPS on ImageNet-1k}
\label{tab:dbscan-best}
\centering
\small
\setlength{\tabcolsep}{4pt}
\begin{tabular}{lrrrr}
\toprule
\multirow{2}{*}{Method} & \multicolumn{2}{c}{NMI} & \multicolumn{2}{c}{ARI} \\
\cmidrule(lr){2-3} \cmidrule(lr){4-5}
 & Best & EPS & Best & EPS \\
\midrule
Supervised       & 0.25 & 0.35 & 0.0007 & 0.37 \\
SimCLR           & 0.23 & 0.21 & 0.0005 & 0.24 \\
BYOL             & 0.31 & 0.10 & 0.0011 & 0.12 \\
Barlow Twins     & 0.21 & 0.16 & 0.0007 & 0.19 \\
VICReg           & 0.34 & 0.16 & 0.0011 & 0.17 \\
DINO             & 0.37 & 0.08 & 0.0016 & 0.09 \\
SwAV             & 0.30 & 0.18 & 0.0011 & 0.20 \\
ReSA             & 0.36 & 0.28 & 0.0016 & 0.30 \\
Hypersolid       & \textbf{0.48} & 0.090 & 0.0030 & 0.11 \\
--- w/o L2       & 0.46 & 0.05 & \textbf{0.0032} & 0.07 \\
\bottomrule
\end{tabular}
\end{table}

DBSCAN depends on selecting a single EPS value, which may not appropriately capture distributions with varying densities. To study this possibility, we also evaluated HDBSCAN, a hierarchical clustering method that can adapt to varying densities. Both versions of Hypersolid (with and without L2 regularization) outperform other methods in terms of NMI, ARI, and number of found clusters by a wide margin, as shown in \cref{tab:clustering-results}. We attribute this result to the low-density gaps between class neighborhoods observed in \cref{fig:energy_2x2}, which may facilitate cluster discrimination and prevent algorithms such as DBSCAN or HDBSCAN from merging distinct clusters. In contrast, other methods exhibit more overlapping clusters, leading to lower clustering quality.

Finally, we also evaluated KMeans with $k=1000$ clusters. Hypersolid again obtained the best NMI and ARI scores; however, the margin of improvement over other methods was smaller than with HDBSCAN. This suggests that while the geometry of other methods is well aligned semantically (as expected), the performance gap between Hypersolid and the other methods when using HDBSCAN or DBSCAN is due to Hypersolid having more pronounced low-density gaps between class neighborhoods.

\begin{table}[!t]
\caption{Clustering quality on ImageNet-1k embeddings using HDBSCAN and KMeans with $k=1000$. For each clustering method, we report normalized mutual information (NMI), adjusted Rand index (ARI), and the number of clusters.}
\label{tab:clustering-results}
\centering
\small
\setlength{\tabcolsep}{6pt}
\begin{tabular}{lccc cc}
\toprule
\multirow{2}{*}{Method} 
& \multicolumn{3}{c}{HDBSCAN} 
& \multicolumn{2}{c}{KMeans ($k=1000$)} \\
\cmidrule(lr){2-4} \cmidrule(lr){5-6}
& NMI & ARI & Clusters 
& NMI & ARI \\
\midrule
Supervised              & 0.1802 & 0.0001 & 330 & 0.6624 & 0.1778 \\
SimCLR                  & 0.0122 & 0.0000 &   2 & 0.5960 & 0.1044 \\
BYOL                    & 0.2776 & 0.0004 & 310 & 0.6275 & 0.1338 \\
Barlow Twins            & 0.0042 & 0.0000 &   2 & 0.6023 & 0.1080 \\
VICReg                  & 0.2998 & 0.0004 & 413 & 0.6353 & 0.1465 \\
DINO                    & 0.3468 & 0.0006 & 408 & 0.6702 & 0.1834 \\
SwAV                    & 0.0061 & 0.0000 &   3 & 0.6405 & 0.1452 \\
ReSA                    & 0.0009 & 0.0000 &   2 & 0.6825 & 0.2038 \\
Hypersolid              & \textbf{0.4826} & \textbf{0.0014} & \textbf{530} & \textbf{0.7045} & \textbf{0.2611} \\
--- w/o L2              & 0.4603 & 0.0013 & 457 & 0.6948 & 0.2374 \\
\bottomrule
\end{tabular}
\end{table}

\subsection{Qualitative Analysis}
As shown in \cref{fig:pca-frontpage}, PCA projections of the hypercolumns suggest object-aligned spatial structure. In multi-object images (e.g., pelicans, horses, cats), the model assigns similar spectral signatures to distinct instances, effectively isolating them from the background and appearing to identify them as belonging to the same semantic class: notice how both pelicans and both cats are shown with a similar color pattern. Notably, foreground objects predominantly map to warmer colors, suggesting that the network utilizes high-variability components to encode salient features. This focus is corroborated by Grad-CAM maps of hypercolumns, which show gradient concentration biased toward foreground entities. Crucially, the feature inversions suggest retention of compositional structure (for instance, a single boat in the sea, rather than many ghost boats floating), but high-frequency details appear to be abstracted away (some are preserved, such as the ring and head patterns of the cats; however, the ship's shape was changed substantially). An explanation of how feature inversion was performed is provided in \cref{sec:APFeatureInversion}.

\newcommand{\frontpagesample}[1]{\includegraphics[width=0.12\linewidth,height=0.12\linewidth]{images_demo/#1.jpg}\includegraphics[width=0.12\linewidth,height=0.12\linewidth]{images/#1_pca_rgb_multilayer.jpg}\includegraphics[width=0.12\linewidth,height=0.12\linewidth]{images/#1_smoothgrad.jpg}\includegraphics[width=0.12\linewidth,height=0.12\linewidth]{images/#1_reconstructed.jpg}}

\begin{figure}[!htbp]
\centering
\frontpagesample{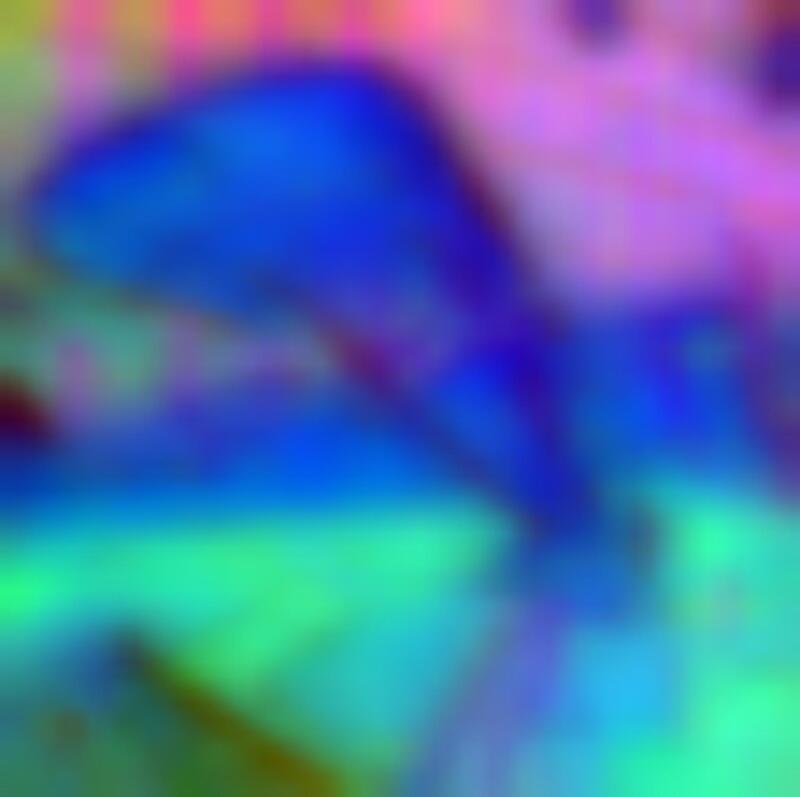}
\frontpagesample{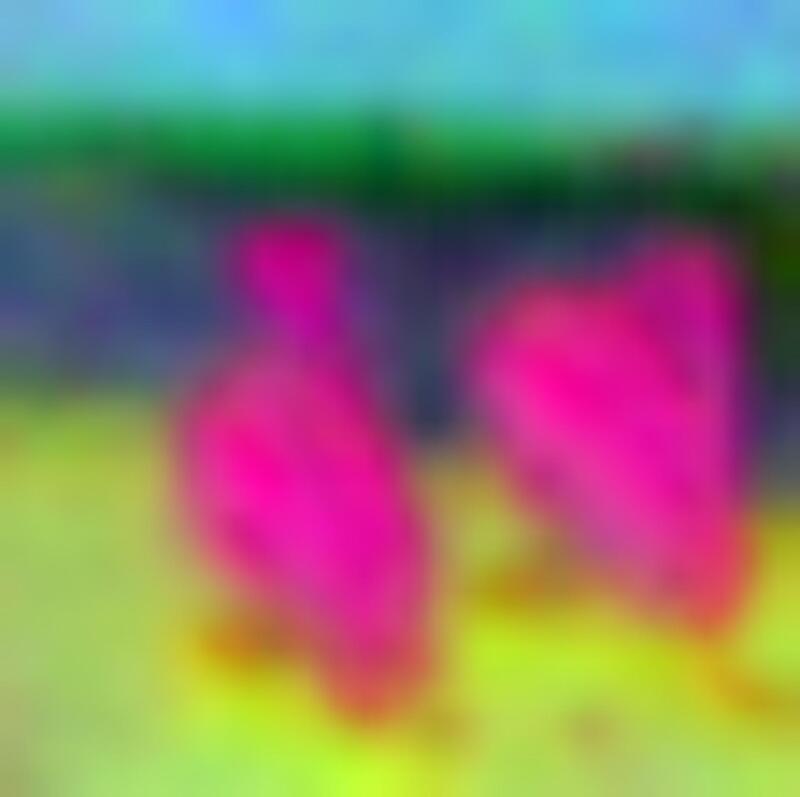}\\
\frontpagesample{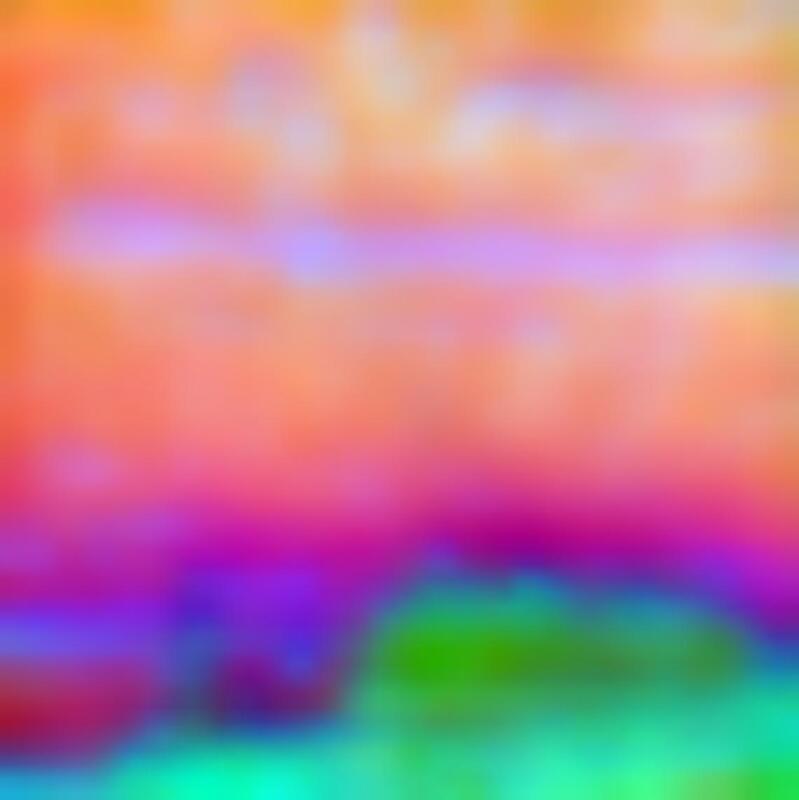}
\frontpagesample{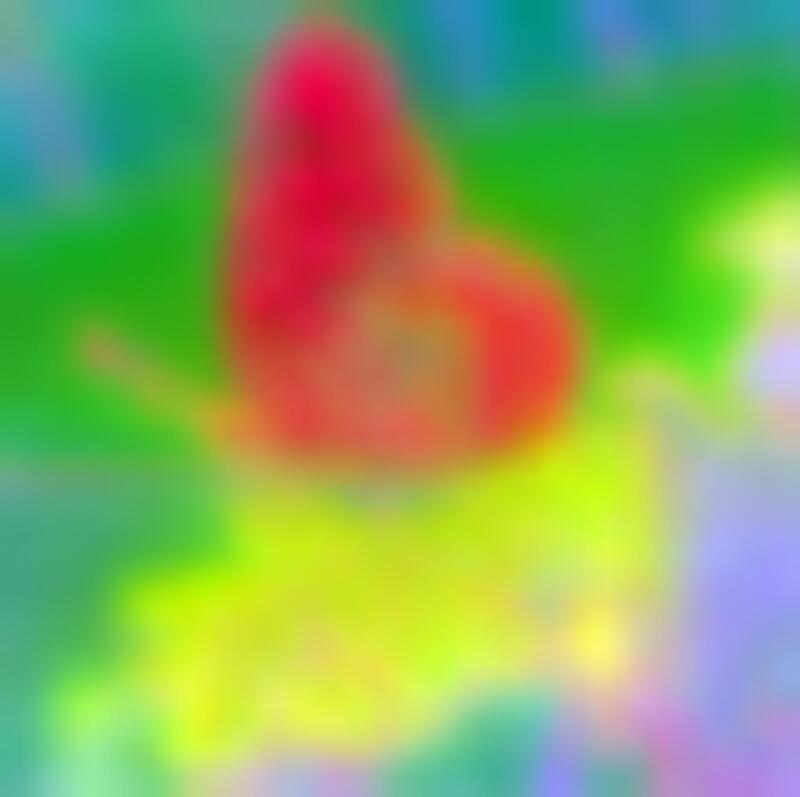}\\
\frontpagesample{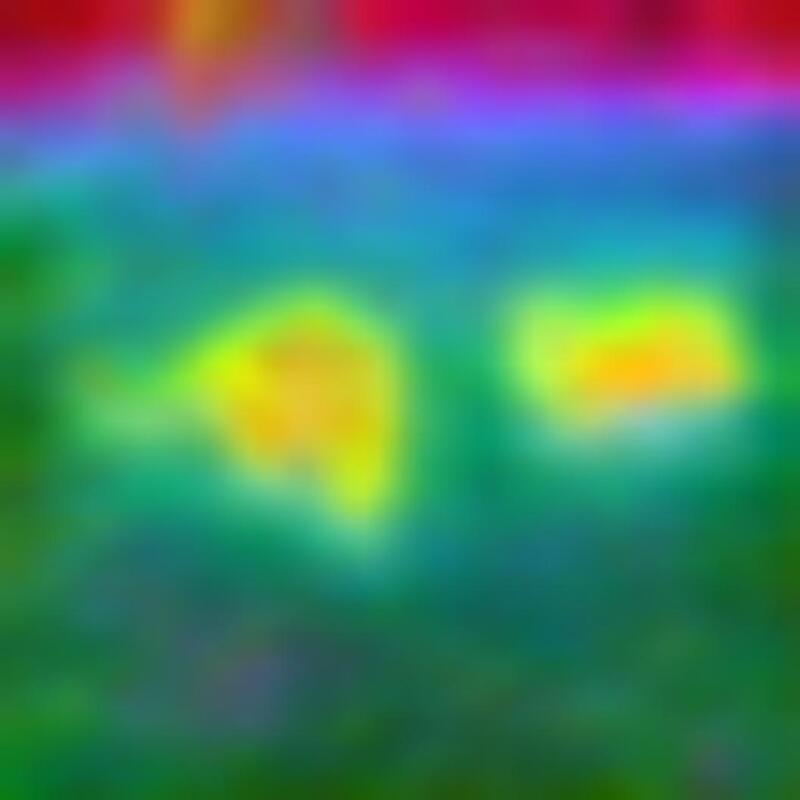}
\frontpagesample{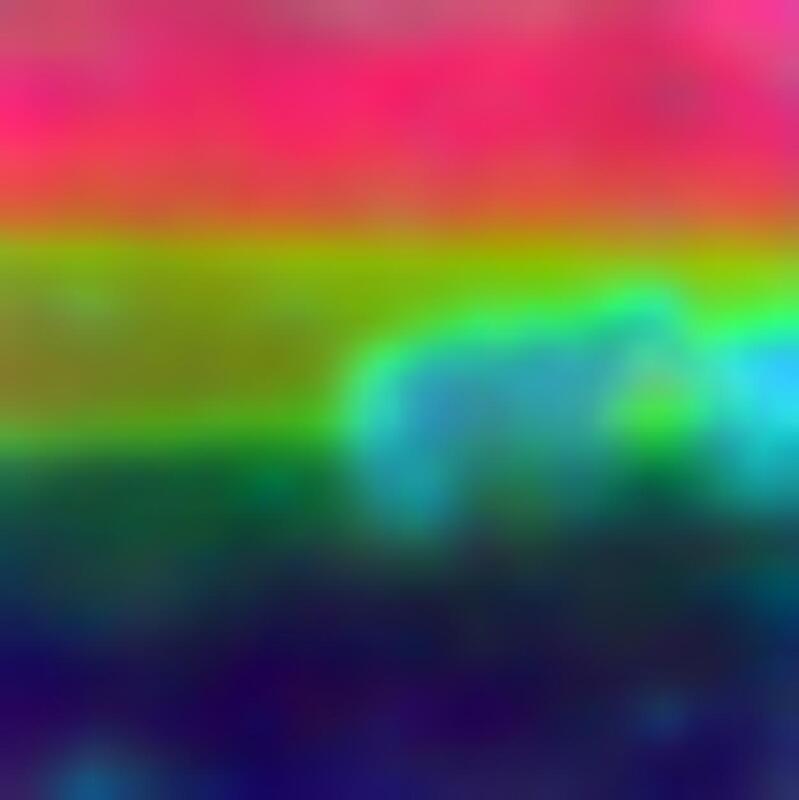}\\
\frontpagesample{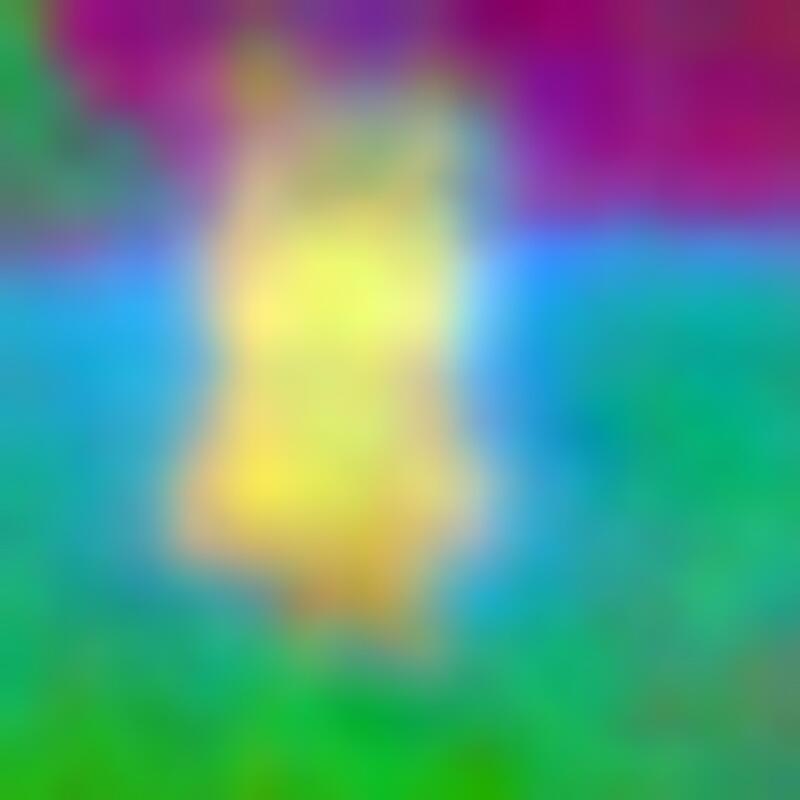}
\frontpagesample{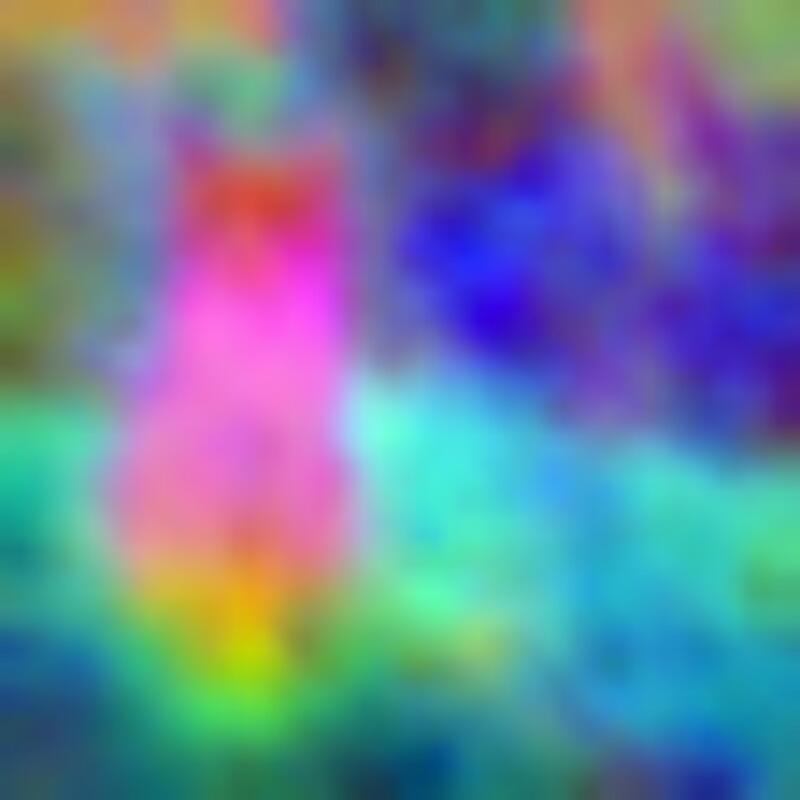}\\
\caption{\textbf{Hypersolid Qualitative Feature Analysis (ResNet-50 on ImageNet-1k).} Left to Right: Input image, hypercolumn PCA, multi-layer Grad-CAM, and gradient-based feature inversion.}
\label{fig:pca-frontpage}
\end{figure}

To contextualize these observations, we compare Hypersolid with DINO. \Cref{fig:dinoPCA} presents the same set of visualizations for a ResNet-50 trained with DINO.

\newcommand{\dinoSample}[1]{\includegraphics[width=0.12\linewidth,height=0.12\linewidth]{images_demo/#1.jpg}\includegraphics[width=0.12\linewidth,height=0.12\linewidth]{images_dino/#1_pca_rgb_multilayer.jpg}\includegraphics[width=0.12\linewidth,height=0.12\linewidth]{images_dino/#1_smoothgrad.jpg}\includegraphics[width=0.12\linewidth,height=0.12\linewidth]{images_dino/#1_reconstructed.jpg} }

\begin{figure}[!htbp]
\centering
\dinoSample{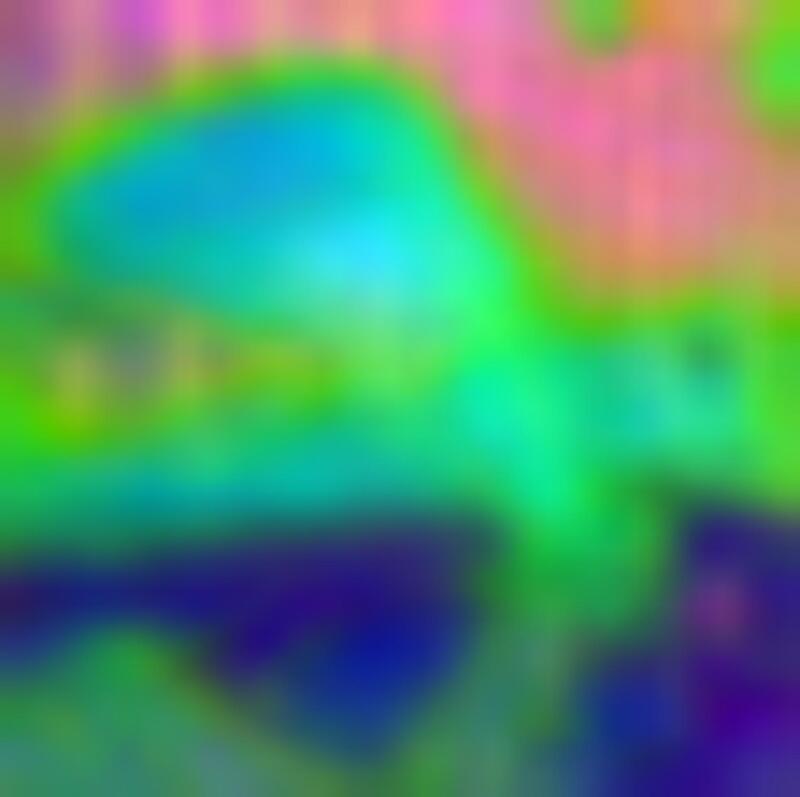}
\dinoSample{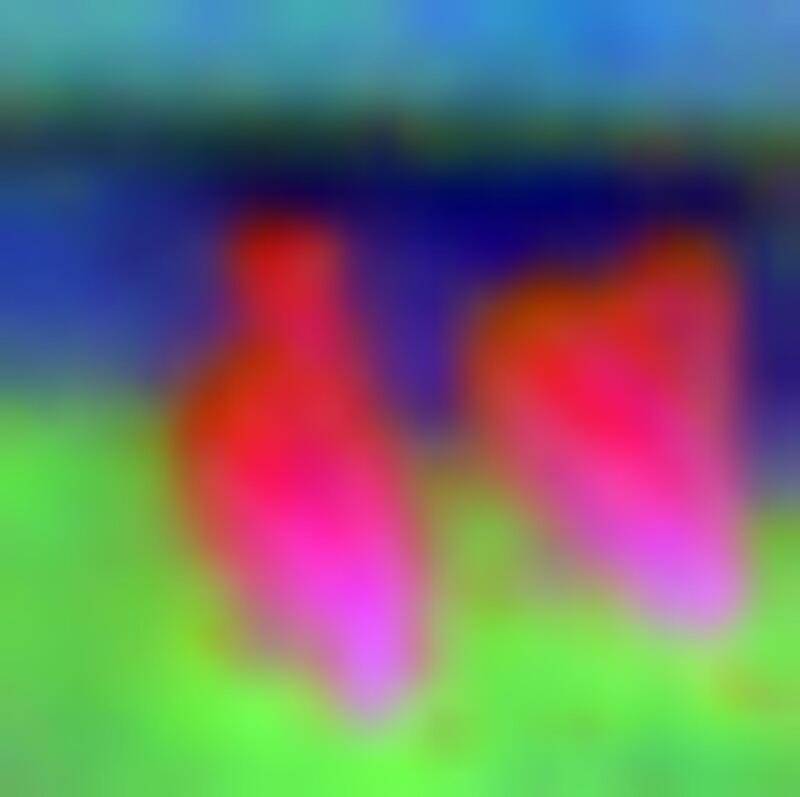}\\
\dinoSample{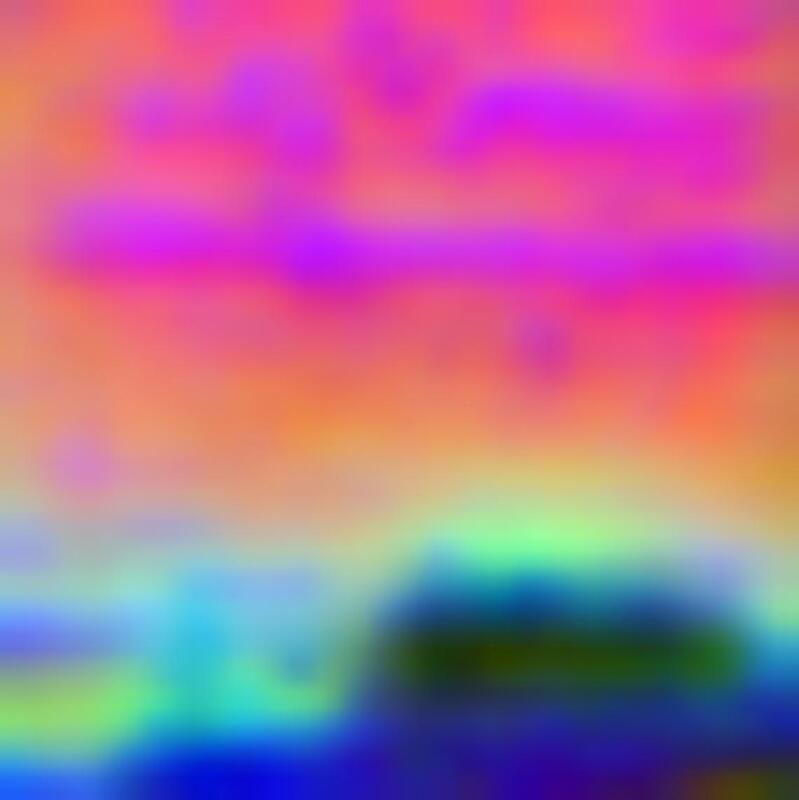}
\dinoSample{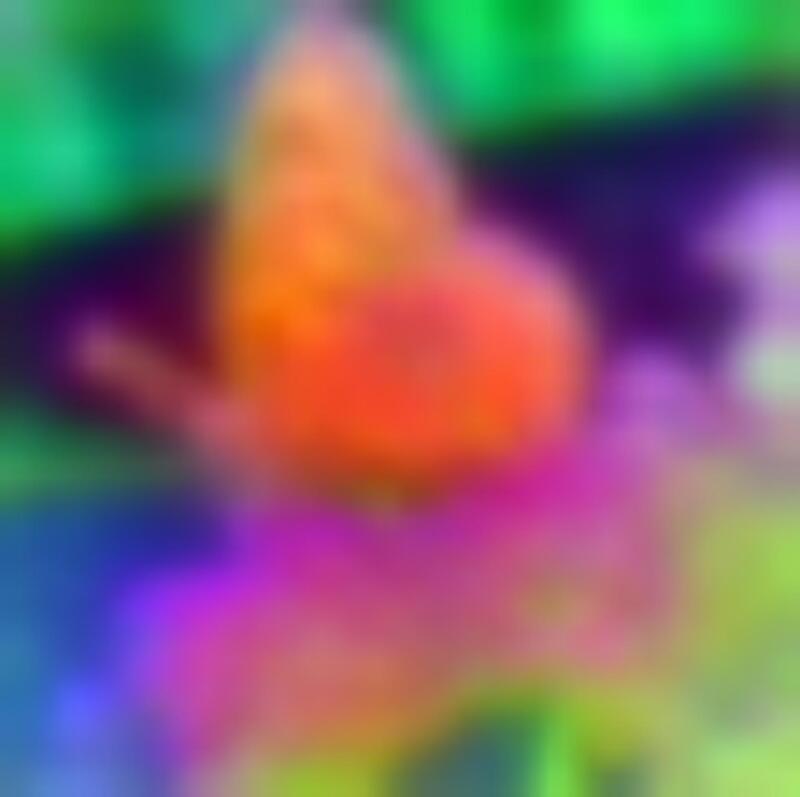}\\
\dinoSample{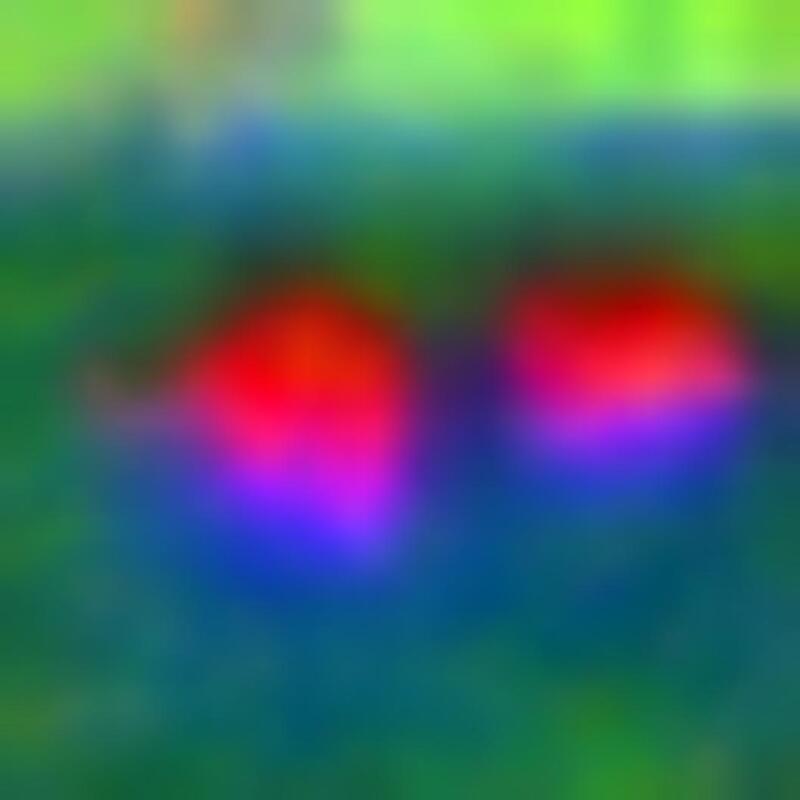}
\dinoSample{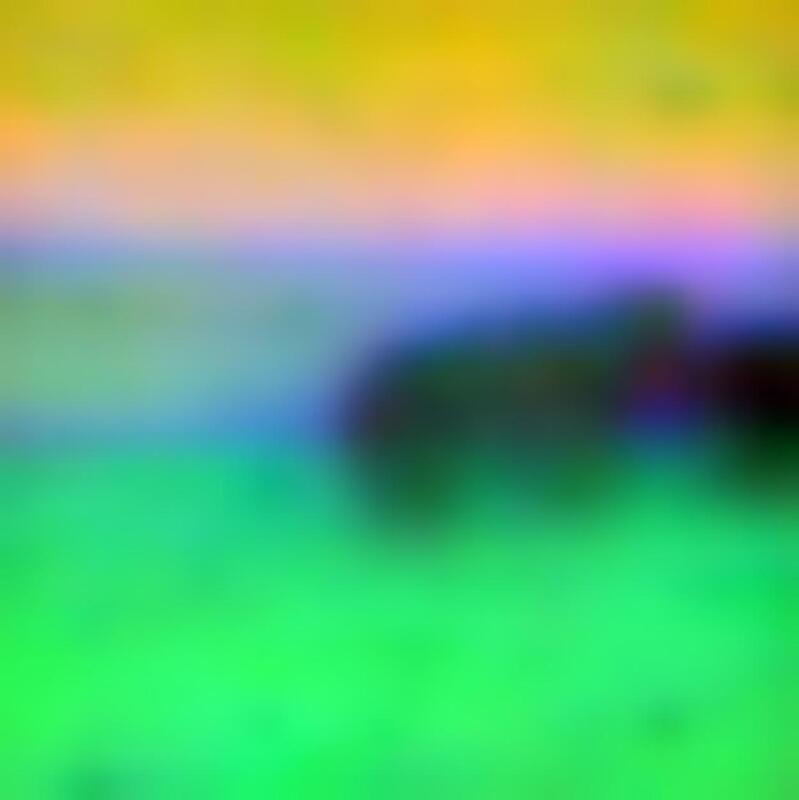}\\
\dinoSample{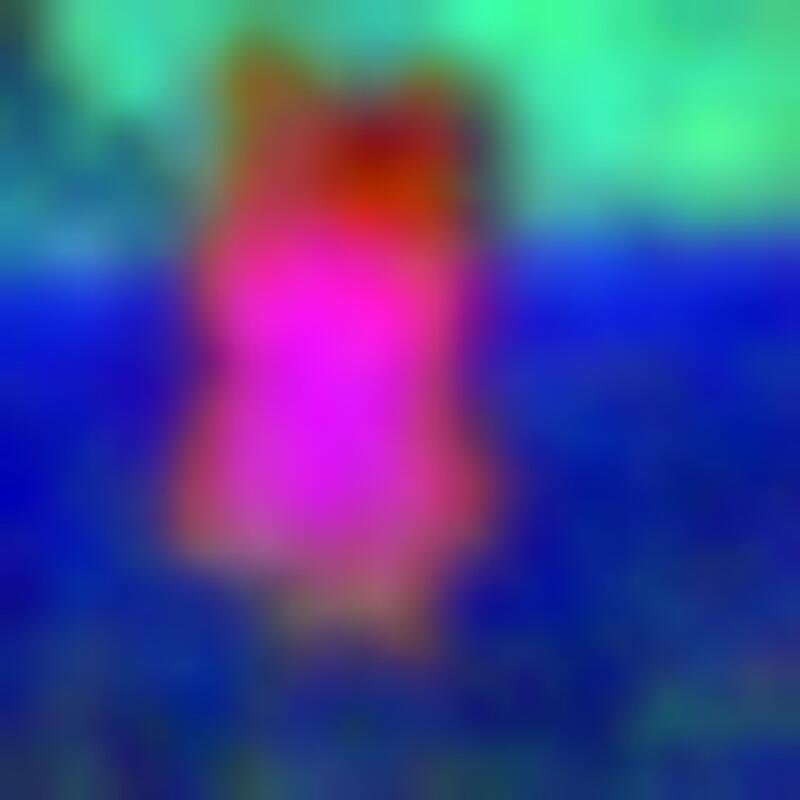}
\dinoSample{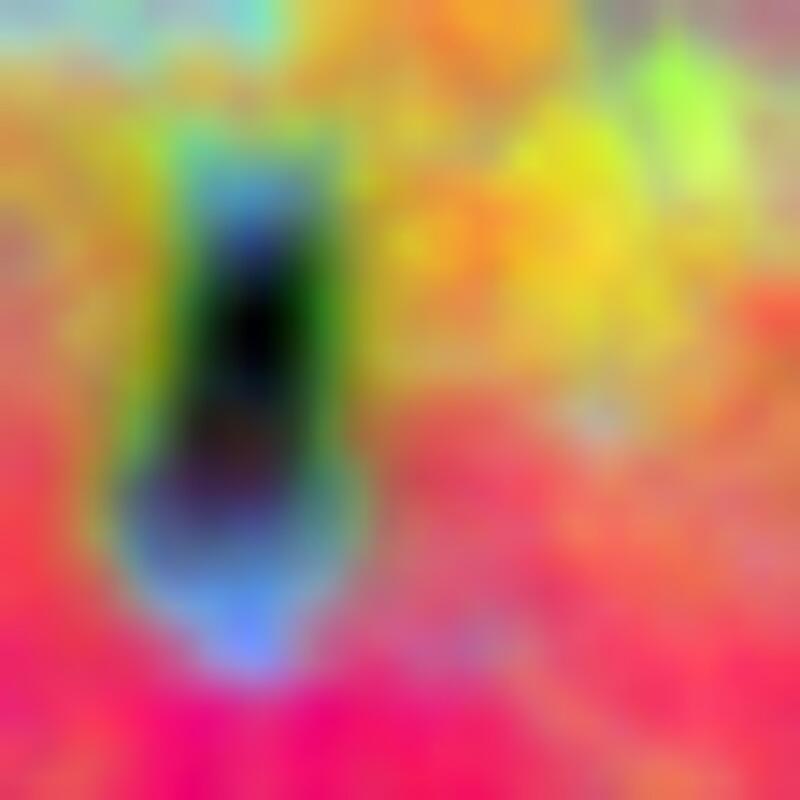}\\
\caption{\textbf{Visualization of learned features by DINO (ResNet-50 trained on ImageNet-1k).} From left to right: PCA projection of hypercolumns, Grad-CAM visualization of all layers, and an image produced using feature inversion.}
\label{fig:dinoPCA}
\end{figure}

Comparing the PCA projections establishes that while both methods successfully isolate objects, they employ different feature subspaces, as evidenced by the distinct color palettes. A key divergence appears in the feature inversions: DINO produces ``collage-like'' reconstructions with higher realism but tends to hallucinate object repetition artifacts (note the boat in the sky and the repeated cats). In contrast, Hypersolid yields ``painterly'' reconstructions that appear to be more aligned with the original scene composition. Interestingly, in the case of the boat reconstruction, Hypersolid reconstructed a red funnel that is not present in the original image, but which is a common feature of boats. This suggests that Hypersolid may be mapping many semantically related inputs into very similar embeddings, while DINO may be encoding much higher diversity. However, this observation is based on a small number of samples and further analysis is needed to confirm this hypothesis.

\section{Hyperparameter Sweeps}\label{sec:hyperparameterSweeps}

To better understand how each component of Hypersolid contributes to its behavior, we isolate the effects of key design choices through controlled sweeps. We first examine the impact of the pooling strategy and $L_2$ normalization on ImageNet-1k with ResNet-50, then present a broader sweep over repulsion, $\alpha$, batch size, learning rate, projector width, and number of views on CIFAR-100 with ResNet-18. Finally, we report observations on the choice of optimizer and augmentation strength.

\subsection{ImageNet-1k and ResNet-50}
We trained a ResNet-50 on ImageNet-1k to compare the effects of mean vs.\ max-pooling and the effect of adding a weak $L_2$ normalization. We used a learning rate of $10^{-3}$, $\alpha = 0.9$, batch size of 128, 1024 projector dimensions, 2 global views + 6 local views, and a normalization weight of $10^{-6}$.

\textbf{Mean vs.\ Max-pooling.} We explored both options for building the alignment target and found that mean-pooling produces faster early learning; however, it quickly stagnates. On the other hand, max-pooling had a slower start but continued improving without stagnating. As a reference, with ResNet-50 trained on ImageNet-1k (both without $L_2$ normalization), we obtained a linear probe accuracy of 38.40\% at the first epoch with mean pooling, while with max-pooling we obtained only 25.52\%. However, after epoch 10, max-pooling surpasses mean-pooling (mean had 47.77\% accuracy and max-pooling 47.80\%). By epoch 48, mean pooling was only at 49.47\%, while the max-pooling version reached 54.78\%.

\textbf{$L_2$ normalization.} We confirmed that a weak $L_2$ normalization can improve the learning process. At a mathematical level, this normalization prevents arbitrary magnitude increase, which was one concern after switching to max-pooling. Moreover, it makes it easier for linear probes to separate points, even if the angles themselves do not change. On ImageNet-1k with a ResNet-50, disabling normalization ($\lambda=0$) resulted in a final linear probe accuracy of 55.95\%, while a weak normalization of $\lambda=10^{-6}$ achieved 61.11\%. We expected this optimization to help only the linear probe; however, it also improved the KNN ($K=200$) probe: without normalization the final accuracy was 47.59\%, whereas with weak normalization the final accuracy at epoch 100 was 50.65\%.

We found it surprising that even a weak penalty of $\lambda=10^{-6}$ can produce global changes in the learned geometry. As shown in \cref{tab:qualityMetrics}, removing the weak L2 regularization increased anisotropy, dimensional correlation, and center vector norm, while decreasing the centroid rank, embedding rank, mean pairwise angle, and sparsity. It increased the sensitivity index slightly. Additionally, it had an effect on our linear probe and K-NN accuracy: under the typical center-crop protocol of ImageNet-1k, removing the L2 penalization improved our linear probe accuracy while decreasing the K-NN accuracy. However, without the center-crop protocol, Hypersolid accuracy increases to $61.10\%$, while Hypersolid without L2 reaches $55.95\%$.

\subsection{CIFAR-100 and ResNet-18}
We trained a ResNet-18 for 200 epochs to isolate the effects of individual hyperparameters (default: $\text{LR}=10^{-3}$, $\alpha=0.9$, batch=128, projections=1024, 6 local views). \Cref{tab:hyperparamsExploration} presents the full sweep results.

\begin{table}[!htbp]
  \caption{\textbf{Hyperparameter Exploration (CIFAR-100).} Top-1 Linear and k-NN ($K=5$) accuracy for a ResNet-18 trained for 200 epochs. Default settings are marked with *.}
  \label{tab:hyperparamsExploration}
  \begin{center}
    \small
    \setlength{\tabcolsep}{4pt}
    \begin{tabular}{@{}c@{\hspace{1.2em}}c@{}}

    \begin{tabular}[t]{@{}lcc@{}}
      \toprule
      Value & Linear Acc & KNN Acc \\
      \midrule

      \multicolumn{3}{@{}l}{\bfseries Repulsion} \\
      All* & \textbf{55.06} & \textbf{51.92} \\
      Neg. Only & 43.46 & 43.42 \\
      Pos. Only & 3.33 & 3.28 \\

      \midrule
      \multicolumn{3}{@{}l}{\bfseries Alpha Value} \\
      0.10 & 52.24 & 46.09 \\
      0.25 & 51.96 & 47.07 \\
      0.50 & 53.64 & 50.07 \\
      0.75 & 55.05 & 52.27 \\
      0.85 & \textbf{55.88} & \textbf{52.81} \\
      0.90* & 55.06 & 51.92 \\
      0.95 & 52.87 & 49.84 \\
      0.975 & 49.21 & 43.89 \\

      \midrule
      \multicolumn{3}{@{}l}{\bfseries Batch Size} \\
      32 & 49.13 & 41.82 \\
      64 & 52.94 & 50.51 \\
      128* & \textbf{55.06} & 51.92 \\
      256 & 54.57 & \textbf{51.99} \\
      512 & 53.97 & 51.08 \\
      1024 & 51.95 & 48.50 \\
      2048 & 48.77 & 45.20 \\
      \bottomrule
    \end{tabular}

    &

    \begin{tabular}[t]{@{}lcc@{}}
      \toprule
      Value & Linear Acc & KNN Acc \\
      \midrule

      \multicolumn{3}{@{}l}{\bfseries LR} \\
      1e-2 & 48.06 & 46.39 \\
      1e-3* & \textbf{55.06} & \textbf{51.92} \\
      1e-4 & 50.73 & 48.04 \\

      \midrule
      \multicolumn{3}{@{}l}{\bfseries Projector Size} \\
      64 & 54.19 & 51.23 \\
      128 & 55.09 & 51.77 \\
      256 & 54.72 & 51.75 \\
      512 & 54.72 & 51.75 \\
      1024* & 55.06 & 51.92 \\
      2048 & 55.65 & \textbf{53.27} \\
      4096 & \textbf{55.96} & 52.02 \\

      \midrule
      \multicolumn{3}{@{}l}{\bfseries Local Views} \\
      2 & 52.83 & 49.28 \\
      4 & 54.95 & 50.89 \\
      6* & 55.06 & 51.92 \\
      8 & 56.22 & 53.61 \\
      10 & 55.50 & 52.90 \\
      12 & 56.16 & 52.72 \\
      14 & 56.38 & 53.56 \\
      30 & \textbf{56.60} & \textbf{53.64} \\
      \bottomrule
    \end{tabular}

    \end{tabular}
  \end{center}
\end{table}

\subsubsection{Repulsion.} Applying repulsion to all pairs yielded the best performance. Restricting repulsion to negative-only pairs caused early instability and lower final accuracy ($-$11.6\% linear probe), while positive-only repulsion failed completely.

\subsubsection{Alpha value.}
We identified a ``goldilocks'' zone for the exclusion radius $\alpha$ between 0.75 and 0.90. Values outside this zone resulted in lower final accuracy. In addition to the CIFAR-100 sweep, we also explored $\alpha$ on ImageNet-100 with ResNet-50, where we found that $\alpha$ values such as $0.25$ or $0.9$ produced lower final accuracy, while $\alpha \in \{0.5, 0.625, 0.75\}$ produced higher and similar final accuracy. This suggests that the optimal $\alpha$ is dataset-dependent.

\subsubsection{Learning Rate.}
With the AdamW optimizer, $\text{LR}=10^{-2}$ leads to faster early accuracy gains, followed by early stagnation. A constant $\text{LR}=10^{-3}$ works well. Smaller values, such as $\text{LR}=10^{-4}$, learn more slowly, but we do not observe learning stagnation.

\subsubsection{Batch Size.}
We swept batch sizes (32--2048) on CIFAR-100. Performance peaked at 128; larger batches slowed learning, while smaller batches caused stalling.
Regarding ViT architectures, we preliminarily found that ViT requires larger batches, as mentioned in \cref{sec:vit_results}.

\subsubsection{Projector Size.}
Performance proved robust to projector width: scaling from 64 to 4096 dimensions yielded $<2\%$ accuracy gain. We settled on 1024 as the optimal efficiency-performance trade-off.

\subsubsection{Number of views.}
Increasing local views yields diminishing returns. While jumping from 2 to 4 views gains 2.1\%, scaling further to 30 views adds only 1.6\% despite the massive computational cost. We found 6 local views to be the optimal efficiency saturation point.

\subsection{Other hyperparameters}

\subsubsection{Optimizer}
During our testing, we found that Hypersolid works better with Adam or AdamW optimizers. With the SGD optimizer, it requires very high momentum (0.99) to learn, and even then it is not as effective as Adam/AdamW. Additionally, we were unable to make Hypersolid learn with the LARS optimizer. We hypothesize that Hypersolid's dynamics could be amplified by the adaptive learning rate of Adam/AdamW, but this effect is not yet fully understood.

\subsubsection{Augmentations} We found that Hypersolid seems to require strong augmentations to learn effectively. Reducing the aggressiveness of the augmentations (e.g., removing multi-crops) resulted in a significant drop in accuracy.

\section{Preliminary Results on Vision Transformers}\label{sec:vit_results}

While our primary analysis focuses on ResNet architectures, we also evaluated the compatibility of our method with Vision Transformers. We trained a ViT-Tiny (patch size 16, image size 224)~\cite{dosovitskiy2020vit} on STL-10 for 100 epochs with a batch size of 64. We maintained the same optimization configuration as our main results (AdamW, learning rate $10^{-3}$, weight decay $10^{-6}$). Using a batch size of 32 did not lead to any meaningful learning.

This model achieved a final linear probe top-1 accuracy of $72.29\%$ and k-NN ($K=5$) accuracy of $67.84\%$. Notably, we observed that ViT backbones require larger batch sizes to converge under our objective. For instance, training with a batch size of 32 resulted in learning stagnation, as accuracy remained below $5\%$. These preliminary results suggest that Hypersolid is not reliant on the inductive biases of convolutional networks. However, fully exploiting the capabilities of Vision Transformers likely requires a distinct hyperparameter exploration, which we did not attempt in this work. Despite this, qualitative analysis reveals a promising property: attention maps from the ViT-Tiny model exhibit distinct object-centric segmentation, often localizing objects effectively. As shown in \cref{fig:tinyVitAttentionPCA}, the network attention often focuses on the foreground object, even with out-of-distribution entities such as the butterfly (not part of STL-10). On the other hand, while the visualization of the PCA projection of the last layer is ``blocky'' (due to the low number of patches in ViT-Tiny), the foreground objects are still distinguishable with colors different from the background.

\newcommand{\tinyvitsamples}[1]{\includegraphics[width=0.10\linewidth]{images_demo/#1.jpg}\includegraphics[width=0.10\linewidth]{images_vit/#1_attn.jpg}\includegraphics[width=0.10\linewidth]{images_vit/#1_pca_rgb.jpg}}

\begin{figure}[!t]
\centering
\tinyvitsamples{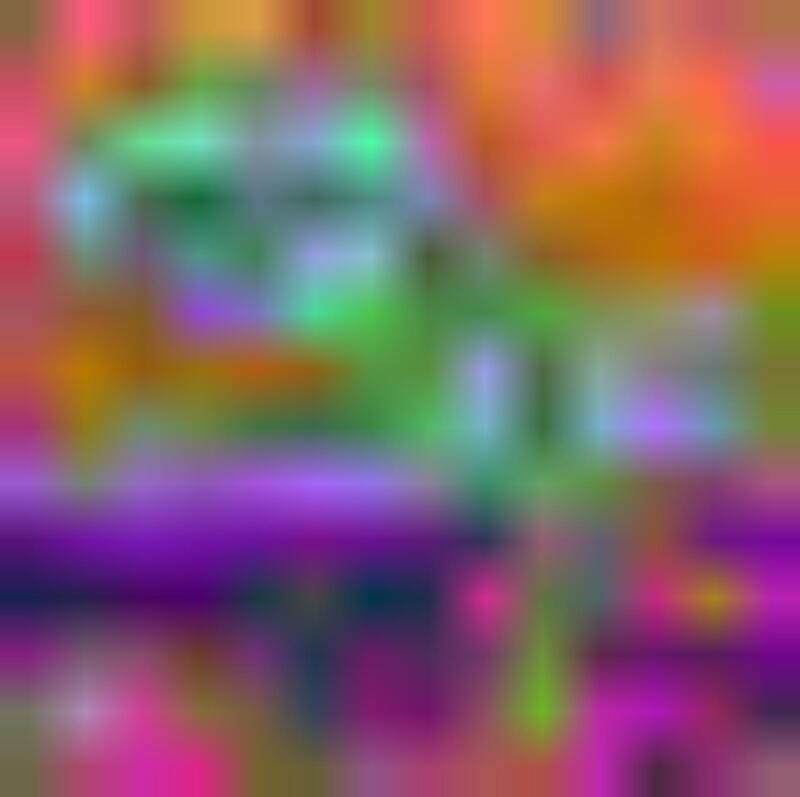}
\tinyvitsamples{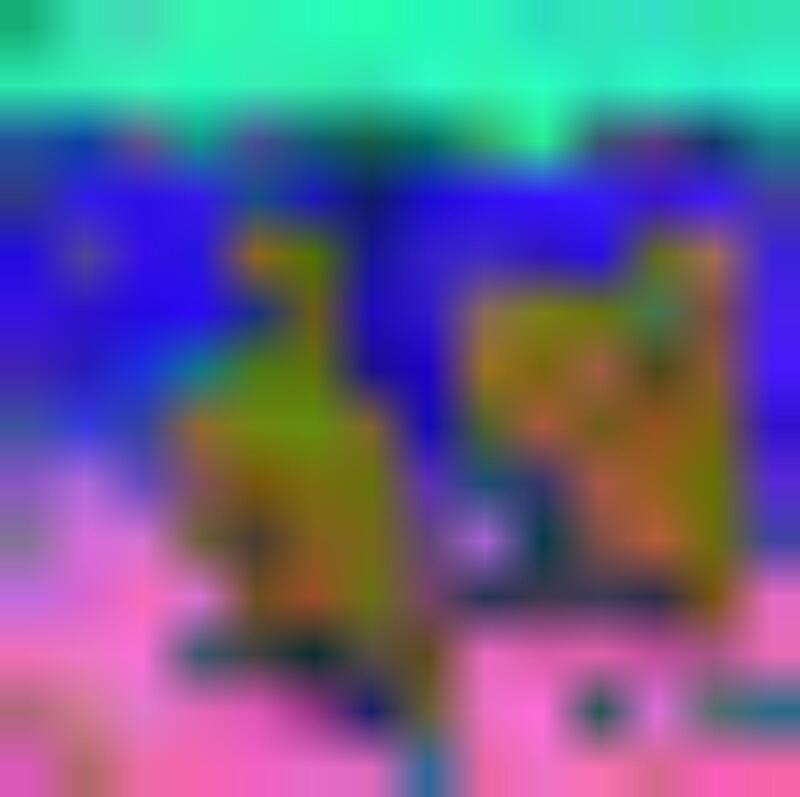}\\
\tinyvitsamples{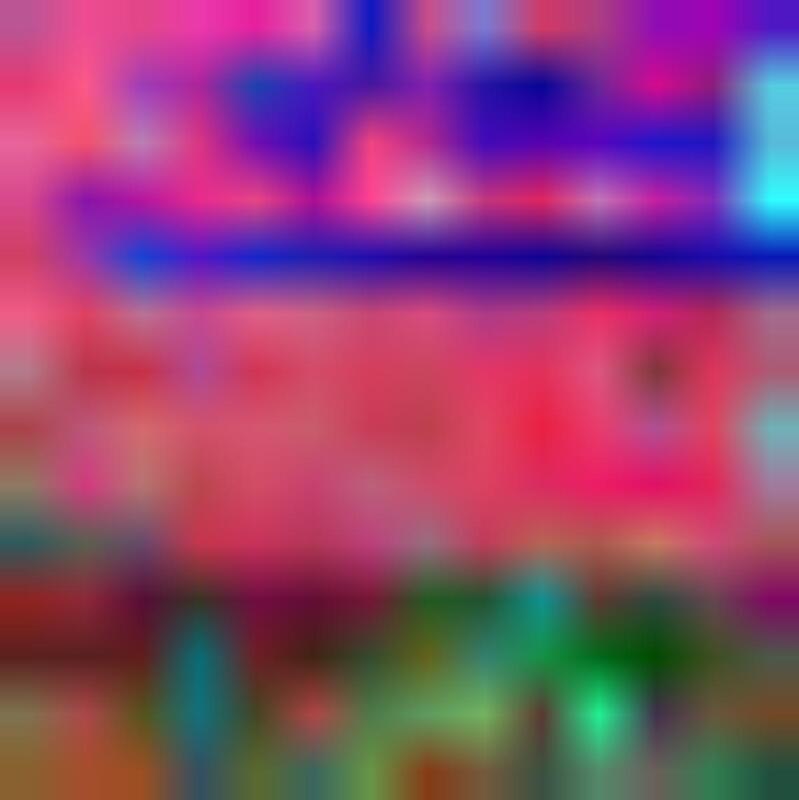}
\tinyvitsamples{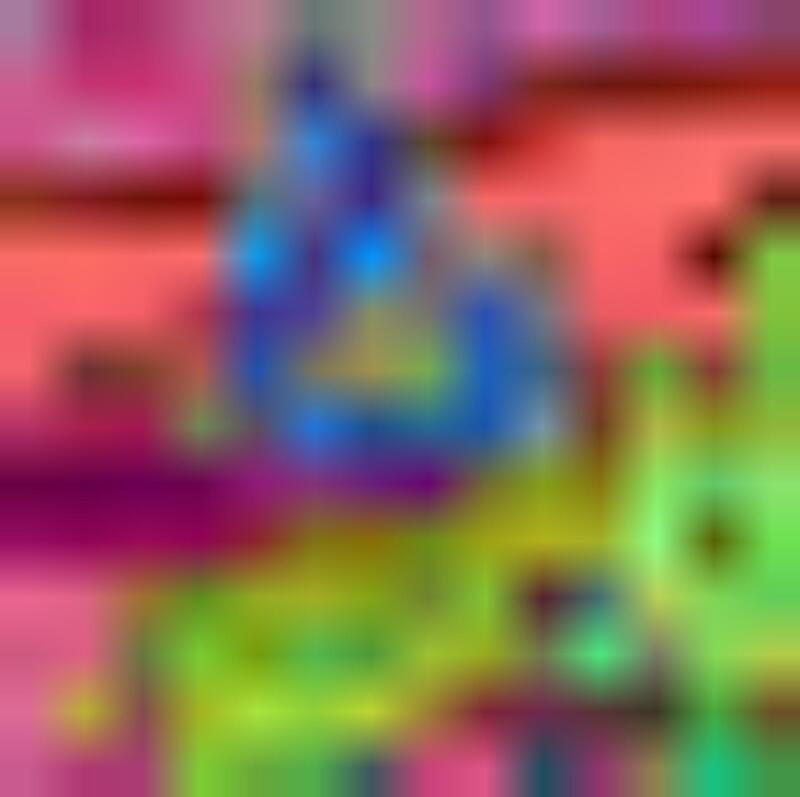}\\
\tinyvitsamples{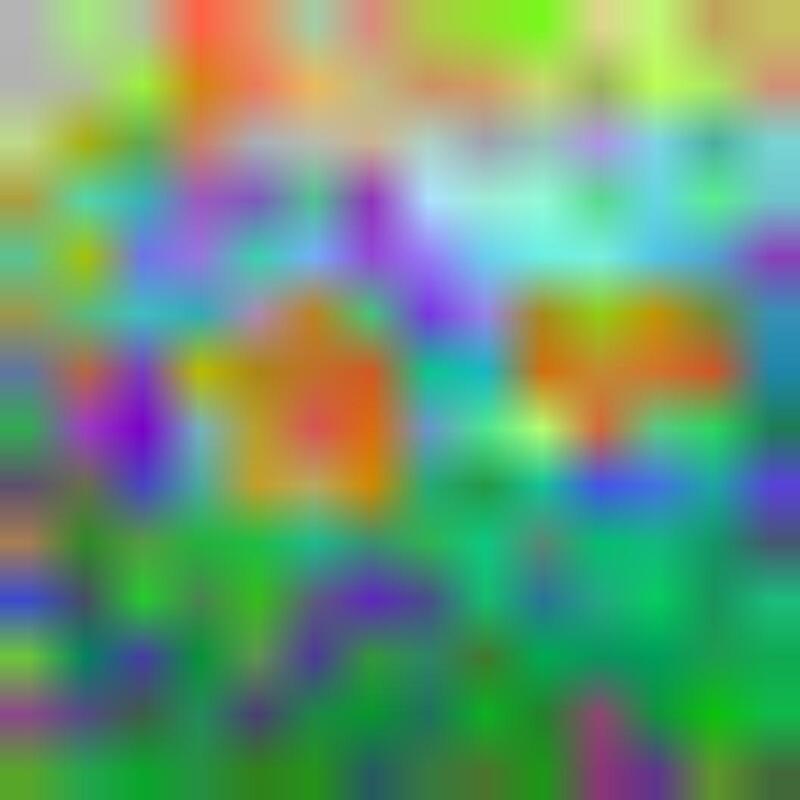}
\tinyvitsamples{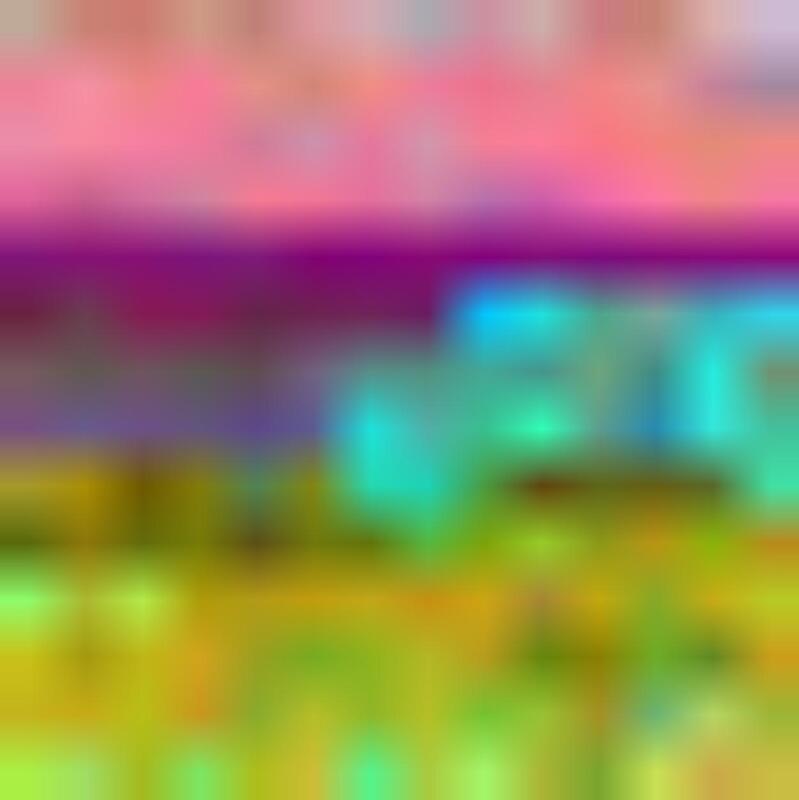}\\
\tinyvitsamples{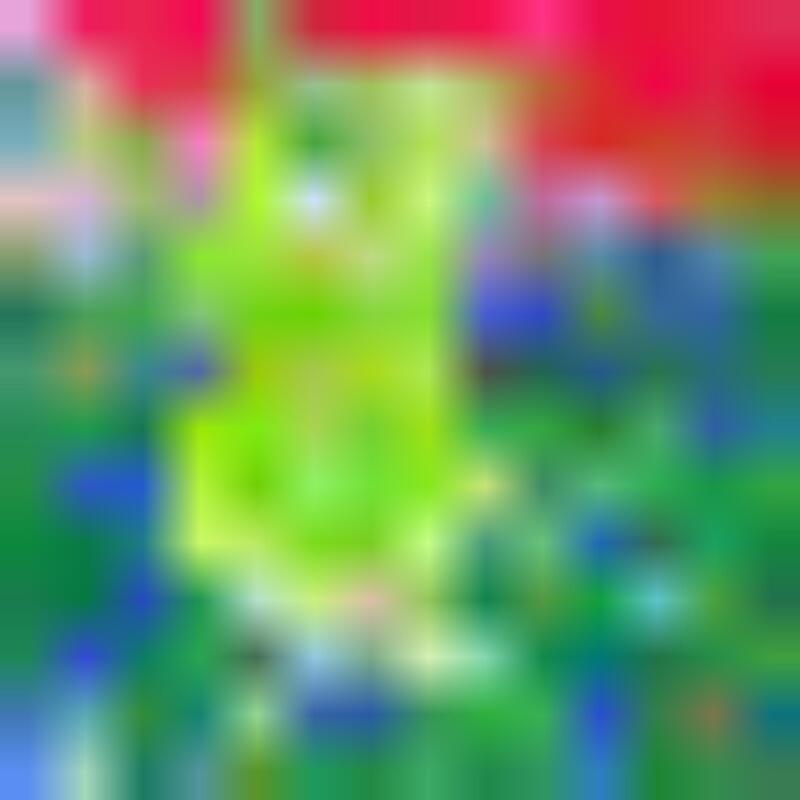} \tinyvitsamples{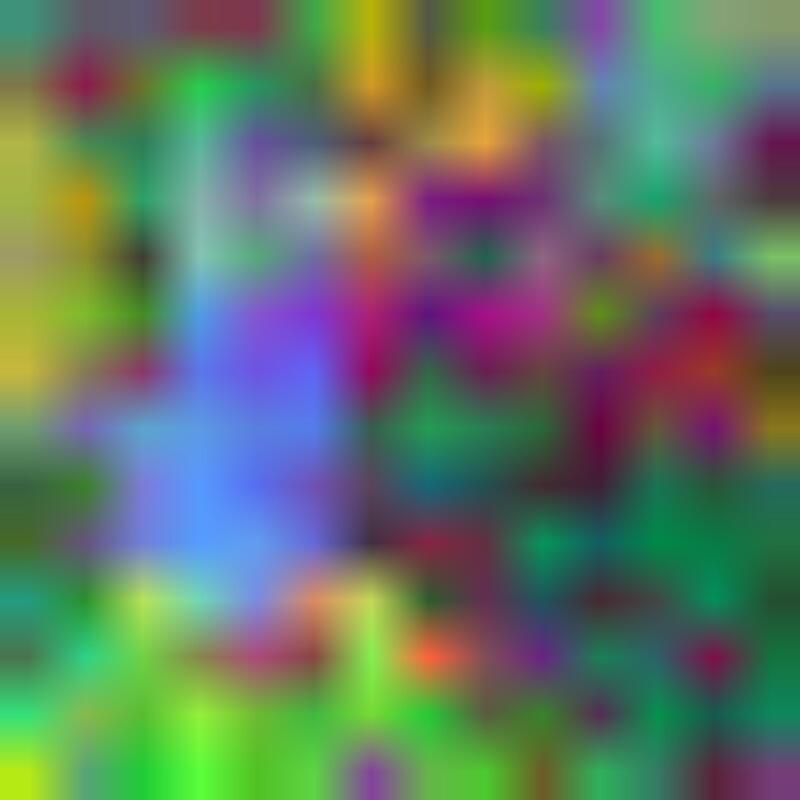}
\caption{\textbf{Qualitative analysis of a ViT-Tiny trained on STL-10.} Despite the limited capacity, data regime (STL-10) and smaller batch size, the model attention still biases toward ``foreground objects''. Due to the lower resolution of the ViT-Tiny patch tokens, the PCA projection is harder to appreciate. Still, some semantic differentiation can be appreciated, such as the pelicans in brown, the two cats in orange or the horses in cyan.}
\label{fig:tinyVitAttentionPCA}
\end{figure}

\section{Limitations and Future Work}

In this work, we prioritized studying the geometric behavior induced by short-range repulsion over large-scale architectural tuning. While our results on ViT-Tiny suggest that the objective generalizes to Transformer architectures and may not depend on convolutional inductive biases, we have not yet conducted the extensive hyperparameter exploration required to establish scaling laws for large-scale ViTs. Additionally, our current implementation utilizes a naive pairwise computation with $O((B \cdot V)^2)$ complexity. However, the short-range nature of the repulsion inherently allows for efficient approximations (e.g., spatial hashing or neighbor lists), which we leave as a primary direction for future optimization.

Given that representations learned by Hypersolid exhibit good clustering properties and the embeddings are highly sparse, we believe that Hypersolid could be a promising candidate for unsupervised clustering and retrieval tasks. We leave the exploration of these applications to future work.

Our experiments are limited by the available compute budget. Most large-scale results are reported from single training runs, and some comparisons rely on public checkpoints or external benchmarks with slight differences in batch sizes and training recipes. Training with multiple seeds and more diverse architectures (particularly vision transformers) remain pending. Despite these limitations, our results demonstrate that short-range repulsion can induce learning a different geometry, which is sparse and classes organize in better defined clusters.

\section{Conclusions}\label{sec:Conclusions}

In this work, we propose to study self-supervised learning through the lens of short-range repulsive geometry. We introduced Hypersolid, a method that operationalizes this geometry through short-range repulsion and feature-union alignment. Our results show that Hypersolid learns an interesting geometry, with a high degree of sparsity and inter-class separation, while maintaining a high degree of intra-class variance. This is useful for unsupervised clustering and fine-grained classification. We hope this work encourages further study of how local repulsive constraints shape the geometry, sparsity, and cluster structure of self-supervised visual representations.

\begin{credits}
\subsubsection{\ackname} We thank the reviewers for their constructive feedback.

\subsubsection{\discintname} The authors have no competing interests to declare that are relevant to the content of this article.
\end{credits}

\bibliographystyle{splncs04}
\bibliography{biblio}

\appendix

\section{Feature Inversion}\label{sec:APFeatureInversion}
To visualize the information retained by the representations, in \cref{fig:pca-frontpage,fig:dinoPCA} we employed a gradient-based feature inversion method. The objective was to synthesize an image $\hat{x}$ such that its embedding $f(\hat{x})$ matches the target embedding $z_\text{target} = f(x)$ of a real image. We optimize the input pixels directly via backpropagation, while keeping the model weights frozen.

We opted to use a multi-scale optimization strategy. The process begins with a low resolution canvas ($28\times28$ pixels) and progressively upsamples to the final resolution ($224\times224$) across 5 scales (0.125, 0.25, 0.5, 0.75 and 1.0). At each transition, the canvas is upsampled using bicubic interpolation.

To generate the image, we minimized the following loss function:
\[
\mathcal{L} = \mathcal{L}_\text{content} + \lambda \mathcal{L}_\text{TV}
\]

where $\mathcal{L}_\text{content}$ minimizes the cosine distance between the current and target embeddings, and $\mathcal{L}_\text{TV}$ suppresses high-frequency noise and checkerboard artifacts, enforcing spatial smoothness:

\[
\mathcal{L}_{TV} = \sum_{i,j} |x_{i+1,j} - x_{i,j}| + |x_{i,j+1} - x_{i,j}|
\]

We used a weight $\lambda = 2$.

Finally, to further improve the resulting images, we applied the following regularization techniques during the optimization loop (Adam optimizer, $\eta=0.05$, 4000 steps per scale):

\begin{enumerate}
    \item Random Jitter: the input image is randomly shifted by up to $\pm 32$ pixels.
    \item Periodic Gaussian Smoothing: Every 50 iterations, we applied a Gaussian blur ($\sigma=0.5$, kernel size $5 \times 5$) to soften away high-frequency noise.
\end{enumerate}

\end{document}